\documentclass{article}
\usepackage{amssymb}

\usepackage[final]{corl_2025} %

\usepackage{graphicx}
\usepackage{amsmath}
\usepackage{amssymb}
\usepackage{booktabs}
\usepackage[dvipsnames]{xcolor}

\usepackage{threeparttable}
\usepackage{tabularx}
\usepackage{array,multirow}

\usepackage{amsfonts,amsthm}
\usepackage{dsfont}
\usepackage{wrapfig}

\usepackage{soul}%
\usepackage[font=small,labelfont=bf]{caption}

\title{FOMO-3D: Using Vision \underbar{Fo}undation \underbar{Mo}dels for Long-Tailed \underbar{3D} Object Detection}

\author{
  Anqi Joyce Yang $^{1,2}$\thanks{Equal contribution.\quad $^\dagger$ Work done at Waabi.}\quad James Tu $^{1,2\ast}$ \\ \textbf{Nikita Dvornik}\footnote{} \quad \textbf{Enxu Li}$^{1,2}$ \textbf{Raquel Urtasun}$^{1,2}$\\
  \vspace{0.01in}\\
	Waabi$^{1}$ \quad University of Toronto$^{2}$ \\
  {\tt\footnotesize{\{jyang, jtu, tli, urtasun\}@waabi.ai}}\vspace{-0.2in}
}

\begin{document}
\maketitle

\begin{abstract}
    In order to navigate complex traffic environments, self-driving vehicles
must recognize many semantic classes pertaining to vulnerable road users or
traffic control devices.  However, many safety-critical objects (\emph{e.g.},
construction worker) appear infrequently in nominal traffic conditions, leading to a
severe shortage of training examples from driving data alone. 
Recent vision foundation models, which are trained on a large corpus of
data, can serve as a good source of external prior knowledge to improve
generalization. We propose FOMO-3D, the first multi-modal 3D detector
to leverage vision foundation models for long-tailed 3D detection. Specifically,
FOMO-3D exploits rich semantic and depth priors from OWLv2 and Metric3Dv2 within
a two-stage detection paradigm that first generates proposals with a
LiDAR-based branch and a novel camera-based branch, and refines them with
attention especially to image features from OWL. Evaluations on real-world
driving data show that using rich priors from vision
foundation models with careful multi-modal fusion designs leads to large gains
for long-tailed 3D detection. Project website is at \url{https://waabi.ai/fomo3d/}.

\end{abstract}

\section{Introduction}

3D object detection is a fundamental task in modern self-driving systems.
State-of-the-art perception models~\cite{yin2021centerpoint,liu2022bevfusion}
can detect and classify common object classes such as \texttt{car} and
\texttt{truck}
reliably well thanks to their frequent occurrences in large-scale urban driving
datasets~\cite{nuscenes,sun2020waymo}. However, these methods struggle to
recognize long-tailed object classes such as \texttt{construction worker} and
\texttt{debris} due to a lack of supervision~\cite{peri2022lt3d}. 
To deploy a self-driving vehicle safely on the road, it is crucial to detect
both common and rare objects well, regardless of their frequency in the real
world. 

Long-tailed class imbalance has been a long-standing challenge in the deep
learning and computer vision community~\cite{zhang2023ltsurvey}. Classic
class-rebalancing methods such as
resampling~\cite{estabrooks2004resample,Kang2020Decoupling} and loss
re-weighting~\cite{lin2017focal,cui2019classbalanced,ren2020balanced} are
popular due to their simplicity. However, they are still restricted to the
original data with few long-tailed examples, resulting in limited success often
at the expense of common class performance~\cite{peri2022lt3d}. On the other
hand, information augmentation
techniques~\cite{zhang2023ltsurvey,yang2020rethink,xiang2020lfme} address class
imbalance by leveraging
external training data or pre-trained
models~\cite{erhan2010pretraining,he2019imagenet}. Inspired by this, we seek
external priors to improve long-tailed 3D detection (LT3D) in self-driving.

\begin{figure*}
    \includegraphics[width=\linewidth]{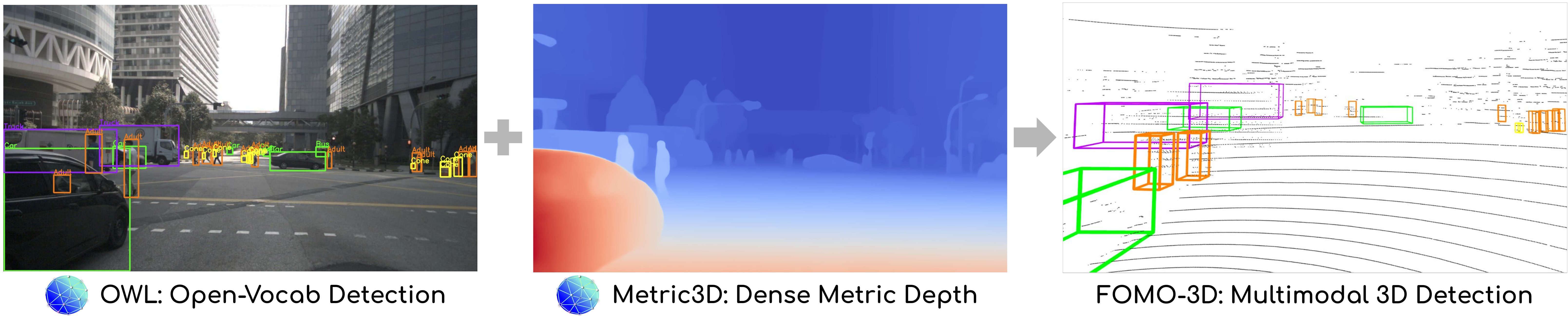} 
    \caption{Vision foundation models OWL (left) and Metric3D (middle) show
    remarkable zero-shot generalization capabilities for 2D object detection and
    monocular depth estimation. Our model FOMO-3D (right) incorporates these
    strong priors along with LiDAR for multi-modal 3D object detection.}
    \label{fig:teaser}   
    \vspace{-5pt}
\end{figure*}

Recent vision foundation models trained on an enormous corpus of internet
images~\cite{radford2021clip,oquab2023dinov2} exhibit remarkable zero-shot
generalization on many vision tasks including
detection~\cite{minderer2022owlvit,minderer2023owlv2}, depth
estimation~\cite{yin2023metric3dv1,hu2024metric3dv2} and
classification~\cite{radford2021clip}. As shown in Fig.~\ref{fig:teaser}, vision
foundation models bring promising prior knowledge for long-tailed detection.
However, they are limited to processing only images, while state-of-the-art 3D object
detectors~\cite{liu2022bevfusion,yan2023cmt,wang2024mv2dfusion} and LT3D
methods~\cite{peri2022lt3d,ma2024mmlf} rely heavily on LiDAR for its 
accurate 3D spatial information. 
Exploiting vision foundation models for 3D
detection therefore requires processing their 2D priors
with 3D sensory inputs.
Fusing LiDAR and camera data is challenging due to their inherently
different modalities. On top of this, 
vision foundation model priors are expressed through different representations, specifically 2D
detections~\cite{minderer2023owlv2}, image features~\cite{radford2021clip,minderer2023owlv2} and
dense depths~\cite{hu2024metric3dv2}. Existing multi-modal fusion
techniques~\cite{liu2022bevfusion,bai2021transfusion,qi2018frustumpointsnet,wang2019frustumconvnet}
cannot accommodate all of these representations simultaneously. Thus, we need a novel fusion method.

To this end, we present FOMO-3D, a multi-modal 3D detector equipped with
novel fusion methods to incorporate different types of 2D image priors 
in conjunction with raw sensor inputs. 
In particular, we adopt OWLv2~\cite{minderer2023owlv2} for zero-shot 2D object
detection and Metric3Dv2~\cite{hu2024metric3dv2} for dense monocular depth
estimation. Built upon a two-stage detection paradigm~\cite{carion2020detr},
our novel camera proposal branch performs early fusion by lifting
OWL detections into 3D, using Metric3D depth and LiDAR to recover accurate 3D
geometry. On the other hand, our proposal refinement performs
feature-level fusion with OWL features to exploit the full semantic and
contextual information in images. To the best of our knowledge, FOMO-3D is
the first to incorporate prior knowledge from foundation models for closed-set
multi-modal 3D detection. We conduct thorough experiments on an urban driving dataset
nuScenes~\cite{nuscenes} and an in-house highway dataset, both with heavily
imbalanced real-world object class distributions. Our evaluations show that
FOMO-3D outperforms existing methods trained on driving datasets
alone, illustrating how powerful foundation model priors, combined with our
careful multi-modal fusion designs, can lead to superior performances on
long-tailed 3D detection.

\section{Related Works}

\textbf{3D Object Detection} is a well-studied problem in self-driving.
Mainstream 3D detectors can be divided into LiDAR-based, camera-based, and
multi-modal detectors. Modern 3D LiDAR detectors~\cite{yin2021centerpoint,
pointpillars,zhou2018voxelnet,chen2023voxelnext} are inspired by image-based 2D
detectors~\cite{ren2015fasterrcnn,liu2016ssd}, and usually adopt convolution in
the Bird's-Eye-View (BEV) or 3D space to process LiDAR point clouds. On the
other hand, camera-based 3D
detectors~\cite{philion2020lift,huang2021bevdet,wang2021fcos3d,jiang2023polarformer,li2022bevformer,jiang2024far3d}
commonly learn depth estimation to lift image information to process in 3D, but
have limited success because monocular depth estimation from images alone is
very challenging. Multi-modal detectors take both LiDAR and camera data as
input.
Early-fusion methods rely on mature detectors for one primary sensor modality,
and fuse the other modality in the input space: they either decorate LiDAR
points with image features to apply LiDAR-based
detection~\cite{vora2020pointpainting,wang2021pointaugment}, or leverage mature
2D detectors and use point clouds to localize 2D detections in
3D~\cite{qi2018frustumpointsnet,wang2019frustumconvnet}. However, these methods
are constrained to the primary sensor modality and suffer from LiDAR sparsity or
2D detection errors. Feature-fusion methods fuse image and LiDAR features in the
BEV space~\cite{liu2022bevfusion, zhao2024simplebevimprovedlidarcamerafusion} or
image and LiDAR space~\cite{yang2022deepinteraction} to decode detections, but
they are not designed to incorporate detection outputs from mature detectors. On
the contrary, late-fusion methods~\cite{peri2022lt3d,ma2024mmlf,pang2020clocs}
directly aggregate detections from a LiDAR detector and a camera detector, but
usually involve sophisticated heuristics to correct detection errors.
Orthogonally, with the success of the two-stage transformer-based 2D detector
DETR~\cite{carion2020detr}, there has been a
trend~\cite{bai2021transfusion,casas2024detra,yan2023cmt,wang2024mv2dfusion} to
represent each object as a query token and perform attention to sensor features
to refine them. Our multi-modal design can be seen as a combination of all these
techniques: we use a two-stage DETR-like framework, where the proposal stage
consists of a mature LiDAR-based detector and a novel camera-focused 3D detector
that performs early fusion on mature 2D detections, and the refinement stage
uses attention and feature-level fusion to refine aggregated multi-modal
proposals.

\textbf{Long-Tailed Perception} has been widely studied in image classifications
and detection.
Class-rebalancing~\cite{estabrooks2004resample,Kang2020Decoupling,lin2017focal,cui2019classbalanced,ren2020balanced}
and information augmentation techniques~\cite{yang2020rethink,xiang2020lfme} are
mainstream solutions to tackle this problem~\cite{zhang2023ltsurvey}.
\cite{peri2022lt3d,ma2024mmlf} are pioneering works to study long-tailed 3D
detection with simple heuristics. \cite{peri2022lt3d} proposes using a single
group-free classifier header for detectors, training with parent object classes
for additional supervision, and a multi-modal filtering (MMF) technique which
removes LiDAR detections that are not in the vicinity of any 3D camera-based
detections. \cite{ma2024mmlf} proposes a multi-modal late fusion (MMLF)
heuristic that alters LiDAR detection scores based on image-space associations
with 2D detections. Both MMF and MMLF assume high recall from LiDAR detections,
which is not true for sparsely observed small and/or distant objects. 
Furthermore, the camera detectors still suffer from limited training examples of
long-tailed objects in driving datasets.
In this work we explore using pre-trained foundation models as an external
prior.

\textbf{Vision foundation models} such as CLIP~\cite{radford2021clip},
DINOv2~\cite{oquab2023dinov2} and EVA-02~\cite{eva02} are trained on an enormous
amount of internet data and contain rich semantic features. Open-vocabulary 2D
object
detectors~\cite{minderer2022owlvit,minderer2023owlv2,zhang2022glipv2,liu2023grounding}
are additionally trained with 2D detection labels and exhibit strong zero-shot
2D detection performance on user-given text prompts. Similarly, monocular depth
foundation
models~\cite{yin2023metric3dv1,hu2024metric3dv2,depth_anything_v1,depth_anything_v2}
based on DINO-v2 and fine-tuned with labelled depth data also have exceptional zero-shot depth
estimations. To leverage vision foundation models, existing camera-based 3D
detectors~\cite{lin2023sparse4dv3,wang2023exploring,liu2023sparsebev} observed
gains by directly using EVA-02 as the image feature extractor backbone, and
existing multi-modal 3D
detectors~\cite{najibi2023unsupervised,etche2024findnprop,zhang2024opensight}
focus on open-set 3D detection and seek to detect novel classes without labels
in the training set. By contrast, we are the first multi-modal method to utilize
vision foundation models in the traditional closed-set 3D detection setting. 
\section{Method}

\begin{figure*}
    \includegraphics[width=\linewidth]{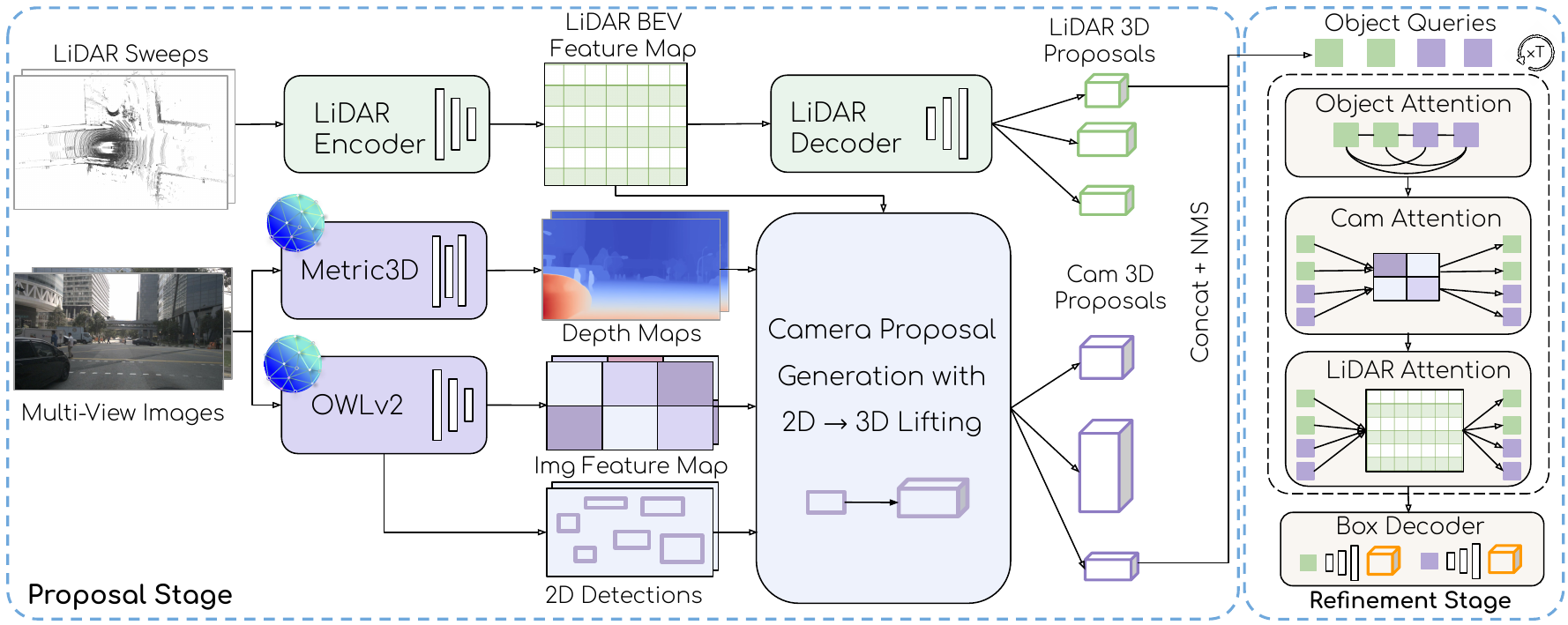} 
    \caption{ Overview of FOMO-3D, which leverages vision foundation models OWL
    and Metric3D, and follows a two-stage paradigm with a multi-modal proposal
    stage and an attention-based refinement stage. }
    \label{fig:overview}   
    \vspace{-5pt}
\end{figure*}

\subsection{Background: Vision Foundation Models and Query-based Object Attention}
\label{sec:background}

To improve long-tailed 3D object detection, our method employs two vision
foundation models: 
OWLv2~\cite{minderer2023owlv2} for 2D object detection and 
Metric3Dv2~\cite{hu2024metric3dv2} for monocular depth estimation. We refer to
these models as OWL and M3D for brevity in the rest of this paper.

\textbf{OWL} is a vision-language model for open-vocabulary object detection.
Given an image and any user-specified text prompts, OWL generates corresponding
2D bounding boxes. Due to an enormous training corpus of over 10 billion
image-text pairs, OWL generalizes exceptionally well to rare objects in traffic
environments. 
OWL consists of two parallel encoders for image and text inputs. The image is
first partitioned into patches of $p \times p$ pixels, and a lightweight
convolutional neural network (CNN) encodes each patch into a transformer token.
The tokens are then processed through a vision transformer~\cite{yuan2021tokens}
to output a final set of tokens $\mathcal{F}_{owl}$.
Each token vector $\mathbf{f}_{owl, i}$ subsequently decodes a 2D bounding box
$\mathbf{b}_{owl,i} = (u_i, v_i, w_i, h_i)$ and a semantic embedding. On the
other hand, the text transformer takes a set of input text prompts $\mathcal{T}$
and encodes each text prompt $t_j$ into the same embedding space. As a result,
each box-prompt pair can be assigned an affinity score based on the similarity
of their respective embeddings.

\textbf{Metric3D} is a monocular metric depth estimation model that takes input
image $\mathbf{I}\in\mathbb{Z}^{H\times W}$ and camera intrinsics
$\mathbf{K}\in\mathbb{R}^{3\times 3}$, and outputs depth map
$\mathbf{D}\in\mathbb{R}^{H\times W}$, expressing the depth of every image pixel
in meters. Pixel-level dense depths are particularly useful in self-driving
datasets, where typically only sparse depths are available via projecting LiDAR
points onto the images. M3D is also trained on a large corpus of data including
real and synthetic depth datasets, and exhibits remarkable zero-shot
generalization performance on outdoor self-driving images.

\textbf{Query-based object detection} first introduced by
DETR~\cite{carion2020detr}, 
represents each object with a learnable object query which is defined as a
feature vector $\mathbf{q}_f \in \mathbb{R}^d$ accompanied by an initial 3D
position $\mathbf{q}_p = (q_x, q_y, q_z)$. Object queries incorporate various
types of information in the scene through transformer attention layers. At a
high level, an attention layer takes as input an object query $\mathbf{q}_f$ and
information $\mathcal{F}$ from the scene (\emph{e.g.}, LiDAR features, image
features, other object queries) and outputs an updated query $\mathbf{q}'_f$ by
attending~\cite{vaswani2017attention} to $\mathcal{F}$. Details of attention can
be found in supp. After $\mathbf{q}_f$ encodes geometry and semantic information
of the object, a multi-layer-perception (MLP) typically processes $\mathbf{q}_f$
to decode a 3D bounding box and the object class. FOMO-3D adopts query-based
attention in the camera proposal branch and in the refinement stage.

\subsection{FOMO-3D: Using Vision Foundation Models for Multi-Modal 3D Detection}
\label{sec:method-overview}
FOMO-3D utilizes 3 types of foundation model outputs: image detections and image
features from OWL, and dense pixel-level depths from M3D. An overall
architecture of how FOMO-3D fuses this information with sensory inputs is shown
in Fig.~\ref{fig:overview}. Built upon a two-stage detection
paradigm~\cite{carion2020detr,casas2024detra,wang2024mv2dfusion},
FOMO-3D first generates detection proposals through two complementary LiDAR and
image branches. The LiDAR branch processes input point clouds to generate
accurate 3D detections. Complementary to LiDAR, the camera branch generates
proposals for rare or small objects that are better distinguished in the image.
Here we lift OWL detections into 3D, utilizing dense M3D depths and a novel
frustum-based fusion module. We then refine the multi-modal proposals through
query-based detection
to incorporate additional information from LiDAR, OWL features, and object
relationships.
Finally, queries are decoded into object classes and BEV bounding boxes $(x, y,
z, l, w, h, \theta)$, denoting centroid $(x,y,z)$, box size $(l,w,h)$ and
heading $\theta$.

\begin{figure*}
    \includegraphics[width=\linewidth]{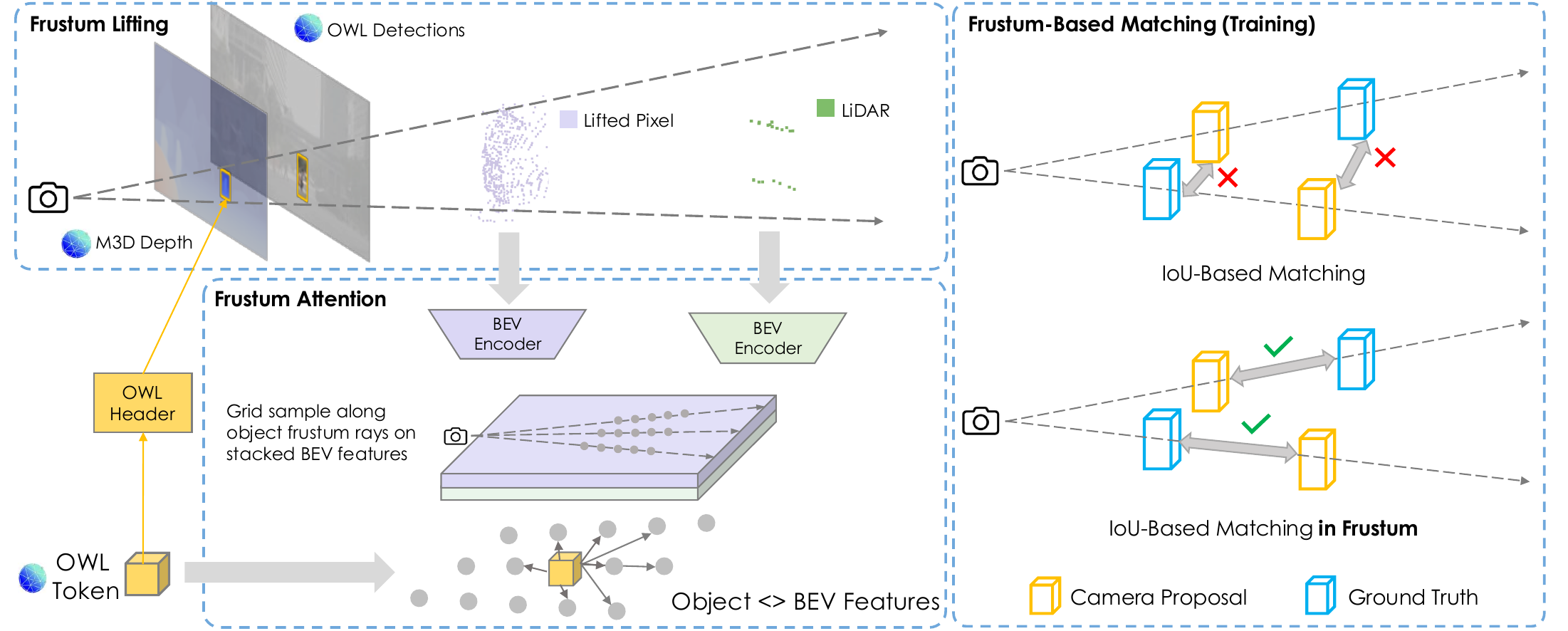} 
    \caption{\textbf{[Left]} Lifting OWL camera proposals to 3D bounding boxes.
    We first unproject pixels inside the camera proposal into 3D using Metric3D
    depths, and then encode the points into a BEV feature map. Each OWL token
    subsequently attends to fused LiDAR and image BEV features sampled along the
    frustum. \textbf{[Right]} During supervision, camera proposals are only
    matched to ground truth boxes inside the object frustum. }
    \label{fig:cam_proposal}   
    \vspace{-5pt}
\end{figure*}

\subsubsection{LiDAR Proposal Generation}
\label{sec:method-proposal-lidar}
Following modern multi-modal 3D detectors, we employ a LiDAR proposal branch to
process point cloud inputs with accurate spatial information.
Specifically, we leverage the single-stage CenterPoint~\cite{yin2021centerpoint}
architecture, which, given a LiDAR point cloud (along with aggregated past few
sweeps), first voxelizes the points into $V_x \times V_y \times V_z$ voxels,
then applies a convolution-based backbone and a feature pyramid network to
obtain a $8\times$ down-sampled BEV feature map $\mathbf{F}_{lidar} \in
\mathbb{R}^{V'_x \times V'_y\times D}$. Finally, following~\cite{peri2022lt3d},
we employ a group-free header design, which uses a single class header and a
single regression header to output multi-class 3D proposals. The class header
decodes feature map $\mathbf{F}_{lidar}$ into a heatmap $\mathbf{H} \in
\mathbb{R}^{V'_x \times V'_y \times C}$, where $C$ is the total number of object
classes, and each $h_{ijk}\in\mathbf{H}$ indicates the probability of an object
of class $k$ present at BEV pixel $(i, j)$. The regression header decodes 3D
bounding box centroids, dimensions, heading angles and velocities for each BEV
coordinate $(i, j)$ in the feature map.
See~\cite{yin2021centerpoint,peri2022lt3d} for more details.

\subsubsection{Camera Proposal Generation}
\label{sec:method-proposal-camera}
In parallel, FOMO-3D employs a camera proposal branch to discover rare and/or
small objects that LiDAR may have missed due to a lack of semantic information
or observability. To this end, OWL's exceptional 2D detection capabilities and
objectness priors motivate us to use image detections directly. 
Thus, we design a camera proposal branch which transforms 2D OWL detections to
3D bounding boxes. At a high level, we first initialize each 2D detection as a
3D object query (Sec.~\ref{sec:background}). We then exploit additional spatial
information from LiDAR and dense depth images, to provide positional and
geometric information to the object queries. Here we introduce a novel
frustum-based attention mechanism to incorporate relevant information within the
3D frustum of each 2D box. A summary is illustrated in
Fig.~\ref{fig:cam_proposal} and we detail our methodology below.

\textbf{Query Initialization.} Following Sec.~\ref{sec:background}, for an input
image $\mathbf{I}$ and text prompts $\mathcal{T}$ pertraining to classes of
interest, we first extract OWL token features and 2D detections
$\{(\mathbf{f}_{owl,i}, \mathbf{b}_{owl,i})\} = \mbox{OWL}(\mathbf{I},
\mathcal{T})$. Please see supp. for details on prompting. Next, we convert OWL
detections into 3D object queries. For each detection $\mathbf{b}_{owl, i}$ with
token feature $\mathbf{f}_{owl,i}$, we initialize the query feature
$\mathbf{q}_{f, i} = \mathbf{f}_{owl, i} + \mbox{PE}([u_i, v_i, d_i]) +
\mbox{PE}(\mathbf{q}_{p,i})$, where $\mbox{PE}$ is the positional encoding
function~\cite{vaswani2017attention}, $d_i = \mathbf{D}(u_i, v_i)$ is the M3D
depth estimate, and the initial 3D query position
\begin{equation}
    \label{eq:unprojection}
    \mathbf{q}_{p, i} = [\mathbf{R}| \mathbf{t}]^{-1} \mathbf{K}^{-1} [u_i, v_i, d_i]^T
\end{equation}
is recovered from unprojecting the 2D box center $(u_i, v_i)$ into 3D using M3D
depth $d_i$, camera intrinsics $\mathbf{K}\in\mathbb{R}^{3\times 3}$, and
extrinsics $[\mathbf{R}| \mathbf{t}]$ with rotation
$\mathbf{R}\in\mathbb{SO}(3)$ and translation $\mathbf{t}\in\mathbb{R}^3$.

\textbf{BEV Feature Construction.} The initial query features $\mathbf{q}_{f,i}$
are derived from the image and lack 3D geometry information. Moreover, the 3D
position of the query may have errors stemming from depth estimation $d_i$. To
address this, our camera proposal branch exploits additional 3D information
expressed explicitly in LiDAR features $\mathbf{F}_{lidar}$ and implicitly
through dense image depths $\mathbf{D}$. Specifically, we generate an
image-based pseudo point cloud as follows: for each image pixel $(u, v)$ inside
any 2D OWL detection, we follow Eq.~\ref{eq:unprojection} to lift it to 3D. Each
lifted point is accompanied by an OWL feature vector at $\mathcal{F}_{owl}(u,
v)$ to preserve semantics information. The image-based feature point cloud is
then voxelized and encoded into a BEV feature map
$\mathbf{F}_{owl,bev}\in\mathbb{R}^{V'_x\times V'_y\times D}$ similarly to
$\mathbf{F}_{lidar}$. We fuse the two BEV feature maps into $\mathbf{F}_{bev}$
by simply concatenating $\mathbf{F}_{bev}(i, j)=\mathbf{F}_{lidar}(i, j) ||
\mathbf{F}_{owl,bev}(i,j)$ at each BEV pixel $(i, j)$.

\textbf{Frustum Attention.} Intuitively, refining the query's 3D position
primarily involves correcting the estimated depth by leveraging information
along the camera frustum. Therefore, following
by~\cite{qi2018frustumpointsnet,wang2019frustumconvnet}, we lift the cropped
image region from the 2D box $\mathbf{b}_i = (u_i, v_i, w_i, h_i)$ to obtain a
3D box frustum, which defines the search range to locate the object in
3D. To further efficiently sample BEV features inside the box frustum, we
construct a mesh grid of $(2N_x + 1)\times (2N_y + 1) \times (2N_z + 1)$
sampling locations around the initial 3D query position $\mathbf{q}_{p,i}$:
\begin{equation}
    \mathcal{P}_i = \left\{\mbox{Unproject}\left(\begin{bmatrix}
        u_i + \frac{p}{2N_x}\cdot w_i \\v_i + \frac{q}{2N_y}\cdot h_i\\d_i + \frac{r}{2N_z}\cdot\delta
    \end{bmatrix}, \mathbf{K}, [\mathbf{R}|\mathbf{t}]\right)\Bigg| \quad\begin{matrix}
        p=-N_x,\dots,N_x \\
        q=-N_y,\dots,N_y \\
        r=-N_z,\dots,N_z\end{matrix}\right\}
\end{equation}
where $\delta$ is a hyperparameter denoting the search distance along the depth
direction, and the unprojection function follows Eq.~\ref{eq:unprojection}. For
each $\mathbf{p}_{\ell} = (x_\ell, y_\ell, z_\ell) \in \mathcal{P}_i$
unprojected from $(u_\ell, v_\ell, d_\ell)$, we retrieve feature from
$\mathbf{F}_{bev}$ via bilinear sampling at BEV location $(x_\ell, y_\ell)$. We
further add spatial information relative to the query point with positional
encoding and derive $\mathbf{g}_\ell = \mathbf{F}_{bev}(x_\ell, y_\ell) +
\mbox{PE}([q_{x, i} - x_\ell, q_{y, i} - y_\ell, q_{z, i} - z_\ell]) +
\mbox{PE}([u_i - u_\ell, v_i - v_\ell, d_i - d_\ell])$. 

Now, for each object query $\mathbf{f}_{q,i}$ representing a 2D OWL detection,
we apply a series of transformer layers to fuse information from the scene.
Specifically, we first apply object self-attention that attends
$\mathbf{f}_{q,i}$ to all object queries $\{\mathbf{f}_{q,i}\}$ in the image to
leverage object relationship cues. Then, we cross-attend with the sampled BEV
features $\{\mathbf{g}_\ell\}$ to effectively incorporate 3D information in the
object frustum. We finally use lightweight MLP layers to decode 3D box
parameters and a class vector $\mathbf{c}_i\in[0,1]^C$ for $C$ classes, where
$c_{ik}$ denotes the probability that object $i$ belongs to class $k$.

At the end of the proposal stage, we derive a set of LiDAR-based 3D proposals
and a set of camera-based 3D proposals. For the multi-camera setting, we apply
the camera proposal model to each camera image independently. To fuse these
multi-modal proposals, we simply concatenate them and deduplicate with
non-maximum suppression (NMS).

\subsubsection{Attention-based Refinement Stage}
\label{sec:method-refinement}
Next, we refine aggregated multi-modal proposals using more general attention
mechanisms to the whole scene. We initialize each proposal as an object query as
follows: the initial 3D position $\mathbf{q}_{p}$ is assigned directly with the
3D centroid of the proposal, and the query feature $\mathbf{q}_f$ is either
$\mathbf{F}_{lidar}(i,j)$ or the final camera query token that decodes the
proposal. Then, each query attends to all object queries, image features
$\mathcal{F}_{owl}$, and LiDAR features $\mathbf{F}_{lidar}$ through a series of
attention layers. We follow~\cite{casas2024detra,wang2024mv2dfusion} for object
self-attention to exploit scene-level object relationship cues, and LiDAR-cross
attention for additional information from the point cloud.
Please see supp. for more details.

To further leverage rich semantic priors from OWL that is essential for object
classification, we perform multi-camera cross-attention to $\mathcal{F}_{owl}$.
Note that different from the camera branch where we attend to lifted image
features in the BEV, here we attend to the 2D features in the image space
directly. Concretely, for each object query, we project its initial 3D position
$\mathbf{q}_p$ onto each camera image $j$ with intrinsics $\mathbf{K}_j$ and
extrinsics $[\mathbf{R}_j | \mathbf{t}_j]$ to obtain image-space coordinates
$(u_{j}, v_{j}) = \mathbf{K}_j[\mathbf{R}_j | \mathbf{t}_j]\mathbf{q}_{p}$.
Then, for each valid projection $(u_{j}, v_{j})$ inside the image, we use a
simple MLP to decode 2D offsets from $\mathbf{q}_f$, sample
$\mathcal{F}_{owl,j}$ at these locations, and perform deformable
attention~\cite{xia2022deformableatt} to obtain an aggregated feature
$\mathbf{h}_j$. Then, we fuse all $\{\mathbf{h}_j\}$ via mean-pooling, and
update the query feature $\mathbf{q}_f$ in the rest of the transformer layer.
Please see supp. for more mathematical details.

\subsubsection{Training and Loss Functions}
\label{sec:method-training}
In accordance with two-stage detection models, we follow a two-stage training
schedule, which trains FOMO-3D's multi-modal proposals branches first, and then
trains the refinement stage with a frozen proposal module. For the LiDAR-branch,
we follow the standard box regression loss from~\cite{yin2021centerpoint} and
group-free heatmap sigmoid focal loss from~\cite{peri2022lt3d}. For the
attention-based camera proposal branch and refinement module, we adopt the
DETR-style set loss function~\cite{carion2020detr,casas2024detra}, with the
modification that for camera proposals, we add a hard constraint in the matching
based on the intuition that the corresponding ground-truth for a camera proposal
(if the ground-truth exists) must fall in the same object frustum. Please see
Fig.~\ref{fig:cam_proposal} for an illustration, and supp. for full details.

\section{Experiments}
We conduct extensive experiments on two real-world datasets to understand the
effectiveness of our approach. In this section, we compare FOMO-3D against
existing works on both datasets, and perform ablations to justify our application
of foundation models.

\begin{table*}[!htp]\centering
    \vspace{-10pt}
    \setlength{\tabcolsep}{10pt}
    \scriptsize
    \begin{tabular}{lccccc}\toprule
    Method & Modality &\texttt{All} &\texttt{Many} &\texttt{Medium} &\texttt{Few} \\\midrule
    FCOS3D~\cite{wang2021fcos3d} & C & 20.9 & 39.0 & 23.3 & 2.9 \\
    BEVFormer~\cite{li2022bevformer} & C & 27.3 & 52.3 & 31.6 & 1.4 \\
    \midrule
    CenterPoint (Group-Free)~\cite{yin2021centerpoint,peri2022lt3d} & L  &39.2 &76.4 &43.1 &3.5 \\
    CenterPoint (Group-Free, w/ Hier.)~\cite{yin2021centerpoint,peri2022lt3d} & L  &40.4 &77.1 &45.1 &4.3 \\
    BEVFusion-L~\cite{liu2022bevfusion} & L  &42.5 &72.5 &48.0 &10.6 \\
    \midrule
    TransFusion~\cite{bai2021transfusion} & L+C &39.8 &73.9 &41.2 &9.8 \\
    BEVFusion~\cite{liu2022bevfusion} & L+C &45.5 &75.5 &52.0 &12.8 \\
    CMT~\cite{yan2023cmt} & L+C &44.4 &\textbf{79.9} &53.0 &4.8 \\
    MMF~\cite{peri2022lt3d} & L+C &43.6 &77.1 &49.0 &9.4 \\
    MMF${^\ast}$ (w/ OWL + M3D)~\cite{peri2022lt3d} & L+C &44.3 &77.7 &46.9 &13.4 \\
    MMLF~\cite{ma2024mmlf} & L+C &\ul{51.4} &\ul{77.9} &\ul{59.4} &\ul{20.0} \\
    \midrule
    FOMO-3D & L+C &\textbf{54.6} &\textbf{79.9} &\textbf{59.6} &\textbf{27.6} \\
    \bottomrule
    \end{tabular}
    \caption{\textbf{[nuScenes] Comparison with SOTA methods} (measured by mAP). L denotes LiDAR, C denotes camera and $^\ast$ is our re-implementation. Our method FOMO-3D improves LT3D performance across all object groups, with a significant 7.6 mAP gain in \texttt{Few} and a 2.0 mAP gain in \texttt{Many}.}
\label{tab:nusc-rarity}
\end{table*}

\begin{figure}
    \includegraphics[width=\linewidth]{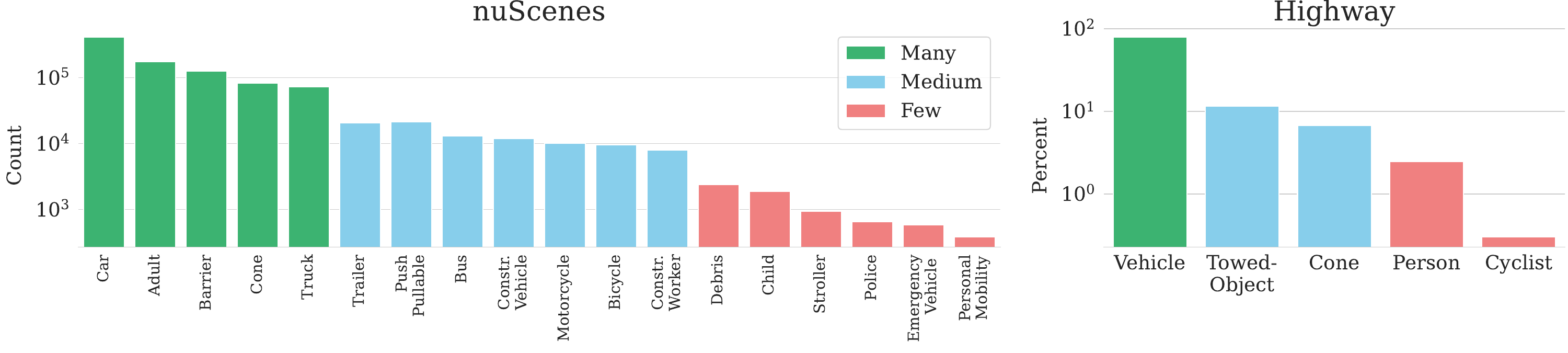} 
    \caption{Real-world class distribution on \emph{nuScenes} and \emph{Highway}. Both
    exhibit severe class imbalances.}
    \label{fig:data_hist}
    \vspace{-5pt}
\end{figure}

\paragraph{Datasets and Metrics.} We experiment with two real-world driving
datasets with diverse LT3D challenges pertaining to urban and highway settings.
First, for the urban setting, following~\cite{peri2022lt3d,ma2024mmlf}, we use
the \emph{nuScenes} dataset~\cite{nuscenes}. \emph{nuScenes} contains 1000
diverse urban scenes %
captured by a top 360$^\circ$ LiDAR and six cameras. We
evaluate on all 18 annotated classes in the validation set, which are divided
into three groups \texttt{Many}, \texttt{Medium} and \texttt{Few} based on commonality.
Following~\cite{peri2022lt3d}, the evaluation range-of-interest
is set to be 50 meters relative to the self-driving vehicle (SDV) for vehicles, 
40 meters for pedestrians, and 30 meters
for movable objects. To further evaluate long-range detection, we use
an in-house \emph{Highway} dataset, which contains over
1700 20-second long training sequences and 400 evaluation sequences, mostly
collected from U.S. highways. We use one front camera for our
experiments, with a detection region of interest of [0, 230] meters
longitudinally and [-50, 50] meters laterally relative to the SDV. We focus on 5
object classes: vehicle, towed object (\emph{e.g.}, trailer), cone, person and
cyclist, where the latter two classes are severely under-represented in the
dataset due to their infrequencies on highways. Fig.~\ref{fig:data_hist} illustrates the class distribution and class
imbalance on these two datasets. Following previous
works~\cite{peri2022lt3d,ma2024mmlf}, we adopt the mean average precision (mAP)
metric over distance thresholds of [0.5, 1, 2, 4] meters. For the aggregated
metrics (\emph{e.g.}, \texttt{Many}, \texttt{Few}), we take the average of mAPs
from relevant object classes.

\begin{table}
    \centering
    \vspace{-3pt}
    \scriptsize
    \setlength{\tabcolsep}{7pt}
    \begin{tabular}{lcccccccccc}\toprule
    Method &Car &Adult &{\color{blue}Truck} &{\color{blue}CV} &{\color{blue}Bicycle} &{\color{blue}MC} &{\color{blue}Child} &{\color{blue}CW} &{\color{blue}Stroller} &{\color{blue}PP} \\\toprule
    MMF~\cite{peri2022lt3d} &\ul{88.5} &86.6 &\ul{63.4} &29.0 &58.5 &68.2 &5.3 &35.8 &31.6 &39.3 \\
    MMLF~\cite{ma2024mmlf} &86.3 &\ul{87.7} &60.6 &\ul{35.3} &\ul{70.0} &\ul{75.9} &\ul{8.8} &\ul{55.9} &\ul{37.7} &\textbf{58.1} \\
    FOMO-3D &\textbf{88.9} &\textbf{90.7} &\textbf{65.1} &\textbf{36.6} &\textbf{72.8} &\textbf{80.2} &\textbf{29.8} &\textbf{60.2} &\textbf{40.1} &\ul{50.0} \\
    \bottomrule
    \end{tabular}
    \caption{\textbf{[nuScenes] Class-specific mAP}. CV = Construction Vehicle. MC = Motorcycle. CW = Construction Worker. PP = Pushable-Pullable. \texttt{Medium} and \texttt{Few} classes are in \color{blue}{blue}.}
    \label{tab:nusc-lca0}
    \vspace{-12pt}
    \end{table}

\paragraph{Implementation details.} We adopt pre-trained
OWL-Large~\cite{minderer2023owlv2} and M3D-Giant~\cite{hu2024metric3dv2} for
best performance. In the camera proposal branch, we set $N_x$, $N_y$, $N_z$,
$\delta$ to 1,1,20,10 for \emph{nuScenes} and increase the depth search range
$N_z=50, \delta=60$ for \emph{Highway}. Frustum-based attention employs object
self-attention followed by BEV feature map attention, for 2 repetitions. We
follow sinusoidal positional encoding according to \cite{vaswani2017attention}.
Our refinement stage employs 2 repetitions of object-image-LiDAR attention
blocks. %
The frustum constraint in
Fig.~\ref{fig:cam_proposal} requires the angle to camera origin between proposal
and ground truth to be less than 0.03 radians.
Please refer to supp. for additional details.

\begin{figure*}
    \includegraphics[width=\linewidth]{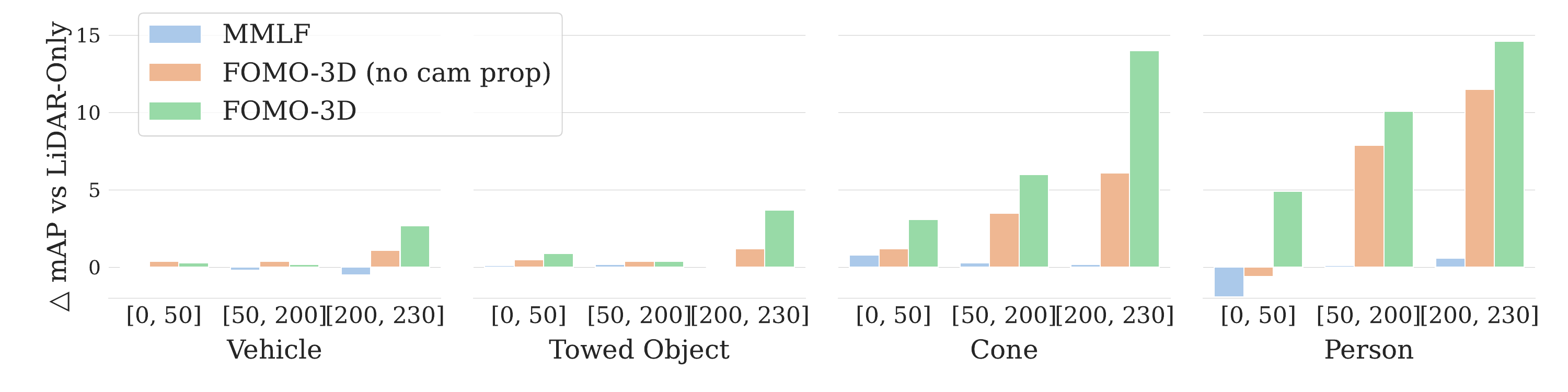} 
    \caption{\textbf{[Highway]} Per-class mAP gains over the base LiDAR-only
    detector, for distance buckets [0, 50], [50, 200] and [200, 230] meters
    relatives to the SDV. \texttt{FOMO-3D (no cam prop)} corresponds to $M_2$ in
    Table~\ref{tab:nusc-rarity-ablation}.}
    \label{fig:waabi-dist}   
\end{figure*}

\begin{figure*}
    \centering
    \includegraphics[width=0.95\linewidth]{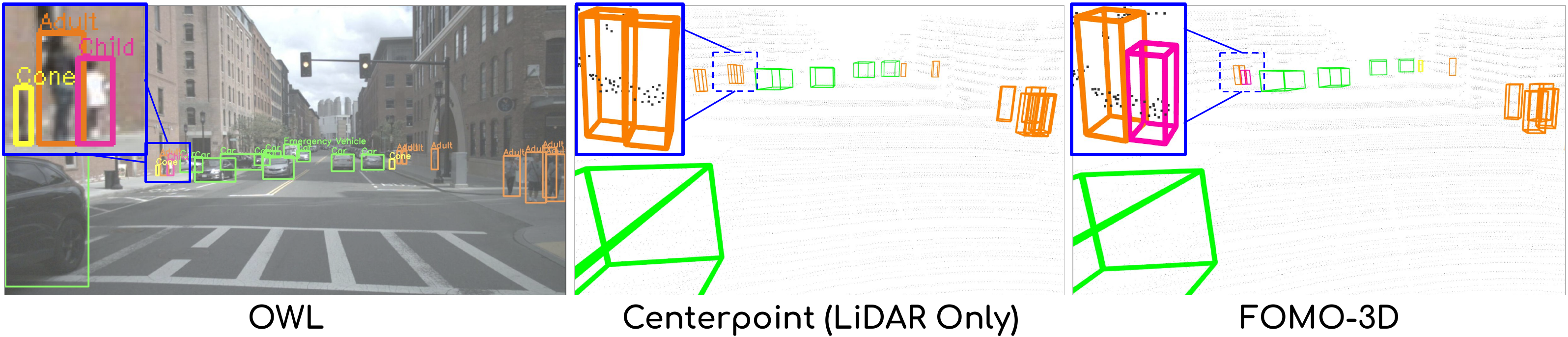} 
    \caption{\textbf{[Qualitative results]} OWL successfully detects the
    \textcolor{magenta}{child} but has a false positive
    \textcolor{Dandelion}{cone}. The LiDAR-only model misclassifies the child as
    an adult. FOMO-3D fuses multi-modal information and foundation model priors
    to generate an accurate 3D bounding box of the \textcolor{magenta}{child},
    while rejecting the false positive \textcolor{Dandelion}{cone}.}
    \label{fig:qual}   
\end{figure*}

\paragraph{Comparison with SOTA methods on \emph{nuScenes}.} We compare with
previous works in the LT3D benchmark~\cite{ma2024mmlf}, including SOTA general
end-to-end learnable 3D detectors, and SOTA LT3D
methods~\cite{peri2022lt3d,ma2024mmlf} that address long-tailed performance
specifically. Table~\ref{tab:nusc-rarity} shows that FOMO-3D outperforms all
existing methods on \emph{every} aggregated object group. Not only does FOMO-3D boost the
mAP of \texttt{Few} from previous best 20.0 to 27.6, it also performs better on
small objects in \texttt{Many}, \emph{e.g.}, cones and adults. Furthermore,
per-class mAP results in Table~\ref{tab:nusc-lca0} show that FOMO-3D outperforms
previous LT3D methods~\cite{peri2022lt3d,ma2024mmlf} for almost every object
class. Fig.~\ref{fig:qual} shows a qualitative result where FOMO-3D successfully
incorporates information from OWL to detect a child.

\begin{wraptable}{r}{6.8cm}
    \vspace{-12pt}
    \scriptsize
    \setlength{\tabcolsep}{3pt}
    \begin{tabular}{lcccccc}\toprule
    Method & Mod. &Vehicle &Towed &Cone &Person &Cyclist \\\midrule
    FOMO-L~\cite{yin2021centerpoint,casas2024detra} & L &86.7 &73.5 &82.6  &58.4 &74.8\\
    MMF$^\ast$~\cite{peri2022lt3d} & L+C &86.3 &73.3 &82.6 &58.3 &75.0 \\
    MMLF$^\ast$~\cite{ma2024mmlf} & L+C &86.6 &73.4 &83.0 &58.3 &76.2 \\\midrule
    FOMO-3D & L+C &\textbf{87.2} &\textbf{74.2} &\textbf{88.9} &\textbf{68.7} &\textbf{79.3} \\
    \bottomrule
    \end{tabular}
    \caption{\textbf{[Highway]} mAP comparisons.}
    \label{tab:waabi-main}
\vspace{-12pt}
\end{wraptable}

\paragraph{Long-range evaluation on \emph{Highway}.} Table~\ref{tab:waabi-main}
evaluates detectors on the long-range \emph{Highway} dataset. FOMO-L
is CenterPoint~\cite{yin2021centerpoint} with
refinement. For fair comparison, we re-implemented the multi-modal filtering
(MMF$^\ast$~\cite{peri2022lt3d}) and late-fusion (MMLF$^\ast$~\cite{ma2024mmlf}) techniques
with the LiDAR detections in Row 1 and camera detections from OWL. For MMF that
relies on 3D camera detection centroids, we lift 2D OWL detection center to 3D
with M3D depths. %
For long-range objects, M3D has bigger depth errors, leading to worse performance for
MMF. In addition, occlusions in busy highway traffic render
image-space association unreliable, resulting in smaller gains from MMLF. By
contrast, FOMO-3D continues to exhibit large gains on all classes, showcasing
its ability to generalize to harder long-range scenarios.
Fig.~\ref{fig:waabi-dist} further illustrates per-class mAP gains against FOMO-L
within three distance buckets [0, 50], [50, 200] and [200,
230] meters. The cyclist
class is not shown due to their absence in certain distance buckets.
FOMO-3D's consistent gains with increasing distances especially for rare
classes highlight its strengths in long-range detection.

\begin{wraptable}{r}{6.9cm}
    \vspace{-12pt}
    \scriptsize
    \setlength{\tabcolsep}{3pt}
    \begin{tabular}{lccccccc}\toprule
        Method & Prop & Refine & Cam Model &\texttt{All} &\texttt{Many} &\texttt{Medium} &\texttt{Few} \\\midrule
        $M_1$ & L & L & - & 42.3 &76.3 &45.1 &10.6 \\
        $M_2$ & L & L+C & OWL &\ul{53.4} &\ul{80.1} &\ul{59.0} &\ul{24.7} \\
        $M_3$ & L+C & L+C & DETR~\cite{carion2020detr} & 47.9 & \ul{80.1} &55.6 & 12.2 \\
        $M_4$ & L+C & L+C & OWL &\textbf{54.6} &\ul{79.9} &\textbf{59.6} &\textbf{27.6} \\
        \bottomrule
        \end{tabular}
    \caption{\textbf{[nuScenes]} Ablations.}
\label{tab:nusc-rarity-ablation}
\vspace{-12pt}
\end{wraptable}

\paragraph{Ablation studies.} To understand the effects of our multi-modal fusion design,
we perform ablations on both datasets. $M_1\rightarrow M_2$ in Table~\ref{tab:nusc-rarity-ablation} and
the positive gains of \texttt{FOMO-3D (no cam props)} in
Fig.~\ref{fig:waabi-dist} show that adding camera attention on top of a
LiDAR-only model leads to big gains due to rich image
semantics. Furthermore, $M_2\rightarrow M_4$ and the gains of \texttt{FOMO-3D}
over \texttt{FOMO-3D (no cam props)} in Fig.~\ref{fig:waabi-dist} show that
camera proposals are indeed complementary and help capture less common
and/or distant objects. Finally,
to justify using OWL for long-tailed
detection, we replace OWL with a 2D detector DETR~\cite{carion2020detr} trained on the
2D detection dataset \emph{nuImages} which follows similar class distributions as \emph{nuScenes}. The
improvement from $M_3$ to $M_4$ in Table~\ref{tab:nusc-rarity-ablation}
shows that FOMO-3D's success on rare classes is indeed attributed to rich
priors from OWL.

\section{Conclusion}
In this paper, we propose FOMO-3D, the first multi-modal 3D detector that
leverages vision foundation models for closed-set 3D object detection.
Specifically, our two-stage model incorporates image-based detections and
features from OWL and monocular metric depths from Metric3D with a novel
camera-based proposal branch and cross-camera-attention in the refinement stage.
On both urban and highway datasets, FOMO-3D outperforms SOTA 3D detectors and
LT3D methods especially on long-tailed classes and long-range objects, and
ablation experiments validate the effectiveness of foundation model priors and
our multi-modal fusion design. With the ability to apply powerful foundation
models to a downstream long-tailed 3D object detection problem, FOMO-3D is a
step towards safer autonomy systems capable of generalizing to rare or unseen
events. 
\clearpage
\section{Limitations}
FOMO-3D employs heavy image foundation models OWL-Large~\cite{minderer2023owlv2}
and Metric3D-Giant~\cite{hu2024metric3dv2}, which are computationally expensive
to run, especially with multiple cameras. As a result, FOMO-3D does not run in
real-time, and is therefore better suited as an offline perception
algorithm~\cite{qi2021offboard,ma2023detzero,yang2023labelformer} which allows a
higher compute budget with applications in auto-labelling. To leverage
foundation model priors for onboard detection, a future direction is to distill
foundation models into smaller models that can run in real-time.

Furthermore, in this work, we take off-the-shelf OWL and Metric3D models and use
their zero-shot results directly. Although the zero-shot detections and depth
estimations are impressive, they are not perfect. One failure mode is that OWL
semantics may be misaligned with respect to specific class definitions in a
particular dataset. For instance, classes that are described similarly in OWL's
internet training corpus (\emph{e.g.}, truck and trailer) may need to be
distinguished in driving datasets. As a result, pre-trained models may confuse
these classes and introduce an information bottleneck compared to processing
input images directly.
To address this limitation, a future direction is to fine-tune pre-trained
foundation models on the downstream dataset, with the goal of improving semantic
alignment while preserving useful priors for long-tailed classes.

\acknowledgments{We sincerely thank our anonymous reviewers for their helpful comments and suggestions. We 
also thank the Waabi team for their invaluable assistance and support.}

\bibliography{ref}  %

\clearpage
\section*{Supplementary Materials}
  \addcontentsline{toc}{section}{Appendix}
  \setcounter{section}{0}
  \renewcommand{\thesection}{\Alph{section}}
  \section{Method Details}
\subsection{Background: Attention}
\label{method:attention}
We use the attention~\cite{vaswani2017attention} mechanism heavily to update
object queries with other information in the scene. Here, we provide
mathematical details of the attention operation.

Attention takes as input a set of $N$ object queries $\mathbf{Q} \in
\mathbb{R}^{N \times d}$, 
a set of $M$ keys $\mathbf{K} \in \mathbb{R}^{M \times r}$
and a set of $M$ values $\mathbf{V} \in \mathbb{R}^{M \times s}$
to output 
\begin{equation}
    \mathbf{A} = \text{softmax}\bigg( \frac{\tilde{\mathbf{Q}}\tilde{\mathbf{K}}^T}{\sqrt{d_k}} \bigg) \tilde{\mathbf{V}} \in \mathbb{R}^{N\times d}.
\end{equation}
$d_k$ is the softmax temperature term, and for brevity $\tilde{\mathbf{Q}} \in
\mathbb{R}^{N\times l}, \tilde{\mathbf{K}}\in \mathbb{R}^{M\times l},
\tilde{\mathbf{V}}\in \mathbb{R}^{M\times d}$ denote linear projections of
$\mathbf{Q}, \mathbf{K}, \mathbf{V}$ with
$\tilde{\mathbf{Q}}=\mathbf{Q}\mathbf{P}_q$, $\tilde{\mathbf{K}} =
\mathbf{K}\mathbf{P}_k$, and $\tilde{\mathbf{V}}=\mathbf{V}\mathbf{P}_v$, and
$\mathbf{P}_q \in \mathbb{R}^{d \times l}, \mathbf{P}_k \in \mathbb{R}^{r \times
l}, \mathbf{P}_v \in \mathbb{R}^{s \times d}$ respectively. Output $\mathbf{A}$
is designed to capture values that are relevant to queries, based on the
similarity between keys and queries. The attention function is general and
object queries can absorb different types of information depending on the choice
of $\mathbf{K}$ and $\mathbf{V}$. In practice, a multi-head attention (MHA)
variation is used for increased expressivity. MHA simply projects $\mathbf{Q},
\mathbf{K}, \mathbf{V}$ with $m$ different projections onto latent dimensions of
sizes $k/m, k/m, v/m$. Then the outputs of attention for each projection are
concatenated together. Under popular transformer nomenclature, we also refer to
queries, keys, and values generally as tokens.

Transformer layers typically use feed-forward networks (FFN) in conjunction with
attention for best results~\cite{vaswani2017attention,casas2024detra}. Following
common transformer architectures, we update object queries following:
\begin{align}
\label{eq:attention-ffn}
    \tilde{\mathbf{A}} &= \text{LN}(\mathbf{Q} + \text{MHA}(\mathbf{Q}, \mathbf{K}, \mathbf{V})) \\ 
    \mathbf{Q} &\leftarrow \text{LN}(\tilde{\mathbf{A}} + \text{FFN}(\tilde{\mathbf{A}})). 
\end{align}
Here LN denotes layer normalization~\cite{ba2016layer}.

\subsection{Attention-based Refinement Stage}
\label{sec:method-refinement}
Following the two-stage detection paradigm, in a second refinement stage FOMO-3D
employs query-based detection to refine all proposals from the first stage.
Different from the frustum-based attentions catered to refining camera proposals
in the object frustum region, here we rely on more general attention mechanisms
to refine the multi-modal proposals in the BEV space.

Refinement starts with initializing object queries. From the LiDAR proposal
branch, each proposal initializes a query using the feature vector from
$\mathbf{F}_{lidar}$ which decoded the box. On the other hand, queries from the
camera proposal branch are taken directly for continued refinement. We then take
the union of these proposals and apply non-maximum suppression (NMS) to remove
duplicates.

Queries are then refined iteratively through a series of transformer decoder
layers. We adopt several kinds of attention layers to model complex correlations
between object queries and multi-sensor inputs in an end-to-end manner, which we
detail below.

\textbf{LiDAR Cross-Attention} adopts the BEV LiDAR feature map
$\mathbf{F}_{lidar}$ and flattens the feature map to obtain a set of key-value
LiDAR feature tokens. Concretely, for $\mathbf{F}_{lidar} \in \mathbb{R}^{V'_x
\times V'_y \times D}$, we flatten it to obtain $V'_x\times V'_y$ LiDAR feature
tokens each with dimension $D$, and we set the values to be the same as the keys
in the attention operation. Different from the LiDAR-based proposal stage where
each proposal is decoded from features from convolution-backbones and subject to
receptive field constraints, the attention mechanism enables object queries to
attend to spatially distant LiDAR tokens. This allows object queries to
incorporate additional information from the input point cloud.

However, since the feature map is a dense representation of the scene, attending
to all LiDAR feature tokens is infeasible to scale. For efficiency reasons, we
employ \emph{deformable attention}~\cite{xia2022deformableatt} - learning a set
of spatial offsets which is added to the object query's location to derive a
sparse set of key-value LiDAR tokens. Specifically, for each object query
$\mathbf{q}_f$ with initial 3D position $\mathbf{q}_p = (p_x, p_y, p_z)$, we
apply a lightweight MLP to $\mathbf{q}_f$ to decode a few 2D spatial offsets
$\{(\delta_{x, i}, \delta_{y, i})\}$, and add to the BEV location $(p_x, p_y)$
to obtain $\{(p_x + \delta_{x, i}, p_y + \delta_{y, i})\}$. We then locate the
associated LiDAR tokens at these BEV locations, and apply the attention
mechanism in Sec.~\ref{method:attention} between each query $\mathbf{q}_f$ and
the sampled LiDAR tokens.

\textbf{Camera Cross-Attention} incorporates information from cameras, which
contains rich semantic cues essential for accurately classifying the object
queries. In this attention layer object queries cross attend to OWL tokens
$\mathcal{F}_{owl}$ which capture semantic and contextual information from the
image. Note that we attend to the 2D camera features here, different from the
previous attention to image features lifted to BEV.

To handle multi-camera inputs, we factorize the attention over each individual
camera, and apply adaptive mean pooling to aggregate features across multiple
cameras. Specifically, for a proposal with initial 3D position $\mathbf{q}_p$
with all valid projections $\{(u_j, v_j)\}$ in camera $j$, we first apply an MLP
to query feature $\mathbf{q}_f$ to obtain a set of 2D offsets $\{(\delta_{u, j,
l}, \delta_{v, j, l})\}$ unique to each $(u_j, v_j)$. Then, we retrieve the set
of OWL tokens at each offset location $\mathcal{F}^{(c_j)}_{owl} =
\{\mathcal{F}_{owl,j}(u_j + \delta_{u,j,l}, v_j + \delta_{v,j,l})\}$, and MHA at
each valid camera yields
\begin{equation}
    \mathbf{A}^{(c_j)} = \text{MHA}(\mathbf{Q}, \mathcal{F}_{owl}^{(c_j)}, \mathcal{F}_{owl}^{(c_j)}) \\
\end{equation}
We next apply mean pooling among all attention matrices across valid cameras:
\begin{equation}
    \mathbf{A} = \text{Mean}(\{\mathbf{A}^{(c_j)}\}). 
\end{equation}
Then we apply $\mathbf{A}$ to update $\mathbf{q}_f$ based on
Eq.~\ref{eq:attention-ffn}.

\textbf{Object Self-Attention} designates object queries as the key-value tokens
 as well. This enables the model to exploit correlations between different
 traffic participants in the scene. The semantic class, object pose, and
 geometry can all be improved via object relationship cues. For example,
 children on the street are often accompanied by adults while parking lot
 vehicles are usually parked in parallel. However, exploiting object
 relationship cues is highly reliant on understanding the relative positions
 between object queries. As positional information has yet to be encoded
 upstream, we also encode each query  with positional embedding
 $\mathbf{q}^{(i)} \leftarrow \mathbf{q}^{(i)} + \mbox{PE}(\mathbf{q}^{(i)}_p)$
 prior to self-attention. Our positional encoding applies sinusoidal positional
 encoding from \cite{vaswani2017attention} followed by a 3-layer MLP.

Our transformer architecture interleaves multiple repetitions of
LiDAR-camera-object attention ``blocks''. This facilitates learning complex
relationships between diverse traffic participants and multi-sensory inputs.
Moreover, after each block, we decode each object query into a detection via a
lightweight MLP. This enables more dense supervision on intermediate outputs of
the transformer, while also allowing refinement to happen in an iterative
manner.

\subsection{Loss Functions}
For the camera-based proposal branch and refinement stage, we apply DETR-style
matching~\cite{carion2020detr} to pair ground-truth labels with detections, and
compute a box regression loss and a classification loss. For the camera-based
proposal, we additionally add a frustum-based hard constraint during matching.
We next detail the matching and losses.

Given $N$ 3D detections and $M$ ground-truths, we first apply Hungarian-based
matching based on a cost matrix $\mathbf{C} \in \mathbb{R}^{N\times M}$ where
$C_{ij}$ indicates the score between the detection $\mathbf{b}_i=[x_i, y_i, z_i,
l_i, w_i, h_i, \theta_i]$ and ground-truth $\mathbf{b}^\ast_j=[x^\ast_j,
y^\ast_j, z^\ast_j, l^\ast_j, w^\ast_j, h^\ast_j, \theta^\ast_j]$. Specifically,
\begin{equation}
    C_{ij} = \lambda_{giou}\mbox{GIoU}(\mathbf{b}_i, \mathbf{b}^\ast_j) + \lambda_{l_2}\|\mbox{Loc}(\mathbf{b}_i) -{Loc}(\mathbf{b}^\ast_j) \|_2,
\end{equation}
where $\mbox{GIoU}$ is the 3D generalized IoU~\cite{rezatofighi2018giou},
$\mbox{Loc}(\mathbf{b}) = [x, y, z, l, w, h]$ is the centroid and dimension of
the 3D box, and $\lambda_{giou}$ and $\lambda_{l_2}$ are hyperparameters.

In addition, for the camera proposal branch, we compute whether the ground-truth
$\mathbf{b}^\ast_j$ falls inside the object frustum of detection $\mathbf{b}_i$
as follows: we first transform both 3D boxes from the reference frame to the
camera frame with the camera extrinsics matrix $[\mathbf{R} | \mathbf{t}]$ to
obtain $\mathbf{b}^{cam}_i$ and ${\mathbf{b}^\ast}^{cam}_j$. Then for any
bounding box $\mathbf{b}$ in the camera frame, we compute the angles between the
box and the camera as $\Phi(\mathbf{b}) = (\mbox{arctan}(x / z), \mbox{arctan}(y
/ z))$, and then we can derive
\begin{equation}
    \mbox{in\_frustum}(i, j) = (\|\Phi(\mathbf{b}^{cam}_i) - \Phi({\mathbf{b}^\ast}^{cam}_j)\|_2 < \alpha_{\phi}) \quad \mbox{and} \quad (\|z^{cam}_i - {z^\ast}^{cam}_j\| < \alpha_z)
\end{equation}
where the first condition compares the camera angles between the detection and
ground-truth and constrains the differences to be under a threshold
$\alpha_\phi$ (set to 0.03 radians in practice), and the second condition
constrains the depths of the two boxes to be no more than $\alpha_z$ apart (set
to 5 meters for \emph{nuScenes} and 30 meters for \emph{Highway}). If
$\mbox{in\_frustum}(i, j)$ is False, we set the corresponding $C_{ij}$ to
\texttt{inf}.

With the computed cost matrix $\mathbf{C}$, we conduct Hungarian Matching to
assign a ground-truth box for each detection. If the associated matching cost is
\texttt{inf}, we discard the matching and the associated detection will remain
unmatched. For each matched detection, ground-truth pair $(\mathbf{b}_i,
\mathbf{b}^\ast_i)$, we compute a box regression loss:
\begin{align}
    \mathcal{L}_{box}(\mathbf{b}_i, \mathbf{b}^\ast_i) &= -\lambda_{giou}\mbox{GIoU}(\mathbf{b}_i, \mathbf{b}^\ast_i) \\
    &+ \lambda_{xyz}(\|x_i - x^\ast_i\| + \|y_i - y^\ast_i\| + \|z_i - z^\ast_i\|) \\
    &+ \lambda_{lwh}(\|l_i - l^\ast_i\| + \|w_i - w^\ast_i\| + \|h_i - h^\ast_i\|)
\end{align}
where $\lambda_{xyz} = 0.2$ and $\lambda_{lwh} = 0.04$ in practice. 

In addition, for each detection $i$ with object class logits $\mathbf{c}_i \in
\mathbb{R}^C$ where $C$ is the total number of object classes, we set the
ground-truth $\mathbf{c}^\ast_i\in\mathbb{R}^C$ as follows: if the detection is
matched to a ground-truth label with class $1 \le k \le C$, then we set
$\mathbf{c}^\ast_i$ to be a one-hot vector with 1 at the $k^\text{th}$ position,
otherwise $\mathbf{c}^\ast_i$ is a zero vector. Then 
\begin{equation}
    \mathcal{L}_{class}(\mathbf{c}_i, \mathbf{c}^\ast_i) = \mbox{SigmoidFocalLoss}(\mathbf{c}_i, \mathbf{c}^\ast_i)
\end{equation}
where the SigmoidFocalLoss first applies sigmoid to the logits $\mathbf{c_i}$
and then uses focal loss~\cite{lin2017focal}.

The final loss is
\begin{equation}
    \mathcal{L} = \frac{1}{N^\ast}\sum_{i}{\mathcal{L}_{box}(\mathbf{b}_i, \mathbf{b}^\ast_i)} + \frac{1}{N}\sum_i{\mathcal{L}_{class}(\mathbf{c}_i, \mathbf{c}^\ast_i)}
\end{equation}
where $N^\ast$ is the number of matched pairs.
  \section{Implementation Details}
\subsection{OWL: Cropping and Prompting}
\label{impl:owl}
OWL~\cite{minderer2023owlv2} usually preprocesses the input image by resizing it
to a square image of fixed dimensions (\emph{e.g.}, 960 by 960 pixels for
OWL-Medium, and 1008 by 1008 pixels for OWL-Large). If the input image is a
rectangular image, it will pad it to a square and then resize. To avoid
information loss from padding, we preprocess our input images preemptively by
cropping the rectangular input image into multiple square crops. For
\emph{nuScenes}, each input image is 1600 pixels by 900 pixels, and we apply two
crops by cropping the leftmost $900 \times 900$ pixels and rightmost $900 \times
900$ pixels of the input image, and run inference with the two crops separately.
We merge the 2D bounding boxes from the two crops with concatenation and
image-based 2D non-maximum suppression with iou threshold 0.85. For camera-based
proposal, each box is associated with the feature embedding from the respective
crop. For camera-attention during refinement, we zero-pad the feature map of
each crop to the original dimension (right-pad for the leftmost crop, and
left-pad for the rightmost crop), stack them, and apply a lightweight
convolutional network to generate a unified feature map of the same size as the
input rectangular image. The convolutional network is:
\begin{verbatim}
    x_proj = Conv2d(in=768*2, out=256, kernel=1)(x)
    x = Conv2d(in=256, out=256, kernel=3)(x_proj)
    x = GroupNorm(num_groups=8, out=256)(x)
    x = GELU(x)
    x = Conv2d(in=256, out=256, kernel=3)(x_proj)
    x = GroupNorm(num_groups=8, out=256)(x)
    x = GELU(x)
    x = x + x_proj
\end{verbatim}

To prompt OWL for 2D detection boxes, we use the following prompts for nuScenes:
\begin{verbatim}
    vehicle.car: ["a car"]
    vehicle.truck: ["a truck"]
    vehicle.trailer: ["a trailer"]
    vehicle.construction: ["a construction vehicle"]
    vehicle.bicycle: ["a bicycle"]
    vehicle.motorcycle: ["a motorcycle"]
    vehicle.bus: ["a bus"]
    vehicle.emergency: ["a police vehicle", "an ambulance"]
    pedestrian.adult: ["a person"]
    pedestrian.child: ["a child"]
    pedestrian.stroller: ["a stroller"]
    pedestrian.construction_worker: ["a construction worker"]
    pedestrian.police_officer: ["a police officer"]
    pedestrian.personal_mobility: ["a scooter", "a wheelchair"]
    movable_object.trafficcone: ["a traffic cone"]
    movable_object.pushable_pullable: ["a dolley", "a wheel barrow", 
                                       "a shopping cart", "a garbage bin"]
\end{verbatim}
If an object class corresponds to multiple prompts, then all boxes associated
with the prompt belong to this object class. Note that we do not have prompts
for barriers and debris because OWL tends to generate a lot of false positives
for these two classes. In addition, we found that ``a person'' is a better
prompt for adult.

OWL outputs each 2D box with an affinity score for each prompt. We take the
argmax of the affinity scores and assigns the object class associated with the
argmax prompt to be the class of the 2D box. Before feeding the list of boxes to
the camera proposal branch, we additionally perform per-prompt score-based
filtering to filter out low confidence 2D boxes. Specifically, we set the
confidence score threshold to 0.2 for car, truck, trailer, bus, construction
vehicle, police vehicle, stroller, scooter and wheel barrow, 0.15 for bicycle,
motorcycle, wheelchair, traffic cone, dolley and shopping cart, 0.1 for
ambulance, person, child, construction worker, police officer, and 0.3 for
garbage bin.

\subsection{Proposal Stage}
For \emph{nuScenes}, we use LiDAR points within distance $[-54, 54]$, $[-54,
54]$ and $[-5, 3]$ meters for the $x$, $y$, $z$ directions respectively. For
\emph{Highway}, we use a longer $x$ range [0, 235] meters along $x$. For
voxelization of both LiDAR and image point clouds, the voxel size is (7.5, 7.5,
20) centimeters along $x$, $y$ and $z$. for \emph{nuScenes} and 15.625cm for
\emph{Highway}.

For the LiDAR-only branch, we follow the group-free wide-512-channel header
CenterPoint implementation from the LT3D codebase~\cite{peri2022lt3d} exactly.
We did not apply the hierarchical heuristic as we did not find it to help with
performance in our experiments.

For the camera-based branch, we provide more details of the model architecture.

\paragraph{Query Initialization and Image Point Cloud Encoding}
The camera proposal branch first applies OWL to the input image with details
specified in Sec.~\ref{impl:owl} to obtain a set of 2D detection boxes with
associated OWL tokens that decode each box. Each OWL token is a feature vector
$\in \mathbb{R}^{1024}$. For each box, we sample the M3D depth map at the 2D box
center with nearest neighbor interpolation to obtain the initial depth. As M3D
can produce degenerate zero depths, we discard the 2D box if the depth is $<
0.5$. The remaining 2D boxes are initialized as queries to lift to 3D.

For each pixel inside any valid 2D detection, we query the M3D depth map and
lift the pixel if the associated M3D depth confidence is $> 0.5$. The lifted
pixels form a 3D pseudo image-based feature point cloud, where each point is
associated with an OWL token of dimension 1024. We then process the lifted point
cloud with a feature encoder and construct a BEV image-based feature map.

The feature encoder consists of a sparse voxelizer followed by a sparse 3D
feature extractor. The voxelizer first voxelizes the 3D point cloud into $V_x
\times V_y \times V_z$ voxels based on the range of interest and voxel sizes
specified above. It applies a linear layer to each point-based feature to reduce
feature dimension from 1024 to 256, and encodes the xyz position with positional
encoding followed by an MLP to add to the reduced feature. To aggregate point
features inside each voxel cell, we mean pool all available point features as
the voxel feature. The resulting sparse voxel feature grid is of dimension $V_x
\times V_y \times V_z \times 256$. We use a sparse voxelizer that only keeps
track of occupied voxels for memory efficiency.

Then, we apply the sparse 3D encoder to the image-based voxel feature grid. To
simplify notation, we use \texttt{Conv3d(in=128, out=128, k=3, s=1, bias=False)}
to denote a 3D sparse convolution layer with input channels 128, output channels
128, kernel size $3 \times 3 \times 3$, stride size $1 \times 1 \times 1$ and no
bias. The sparse 3D encoder consists of \texttt{Conv3d(in=256, out=128, k=3,
s=1, bias=False)}, followed by BatchNorm (BN), ReLU, and two repetitions of
\texttt{Conv3d(in=128, out=128, k=3, s=1, bias=False)} $\rightarrow$ BN
$\rightarrow$ ReLU $\rightarrow$ \texttt{Conv3d(in=128, out=128, k=3, s=1,
bias=False)} $\rightarrow$ BN. For each voxel cell, we further apply a linear
layer to the feature with input and output dim 128 and bias=True. Next, we
squash the 3D voxel grid to BEV by mean pooling the features along the z
dimension. The BEV feature map goes through a dense block of
\texttt{Conv2d(in=128, out=128, k=3, s=1, bias=False)} $\rightarrow$ BN
$\rightarrow$ \texttt{Conv2d(in=128, out=128, k=3, s=1, bias=False)}
$\rightarrow$ BN $\rightarrow$ ReLU, and then a sequence of
\texttt{Conv2d(in=128, out=256, k=1, s=1, bias=False)} $\rightarrow$ BN
$\rightarrow$ ReLU to finally decode a BEV feature map of size $V_x \times V_y
\times 256$.

\paragraph{Frustum-based Attention}
For each feature initialized with OWL tokens $\in \mathbb{R}^{1024}$ and
positional encodings over $xyz$ and $uvd$, we first apply a lightweight MLP with
\texttt{Linear(in=1024, out=512, bias=False)} $\rightarrow$ LayerNorm (LN)
$\rightarrow$ ReLU $\rightarrow$ \texttt{Linear(in=512, out=512, bias=False)}
$\rightarrow$ LN $\rightarrow$ ReLU $\rightarrow$ \texttt{Linear(in=512,
out=512, bias=False)} to reduce feature dimension to 512. To include
classification information from OWL, we also encode the OWL class logits that
represent the affinity score between the detection and all prompts. We encode
the OWL class logits with an MLP that has a similar architecture to the feature
dimension reduction MLP except the input dimension of the very first linear
layer is the number of OWL prompts. We add these two encodings as the updated
query feature to be used in the attention operations.

To efficiently sample features from the merged LiDAR and image BEV feature map,
we sample along frustum rays as detailed in the main paper. We then apply two
repetitions of attention blocks, where each block starts with an
object-to-object self-attention, and an object-to-sampled-BEV-features
cross-attention. The attention layer~\cite{vaswani2017attention} follows
Sec.~\ref{method:attention} and is a standard pytorch
\texttt{TransformerEncoderLayer} layer with the attention operation followed by
two FFNs. Both the object-to-object self-attention and the object-to-BEV-feature
cross-attention layers are a single transformer encoder block with 8 attention
heads, input and FFN feature dimension 512, bias=False and dropout=0.1.

At the end of each attention block, we use a lightweight MLP to decode the 3D
proposal box parameters. The box decoder is a sequence of \texttt{Linear(in=512,
out=256, bias=False)} $\rightarrow$ LN $\rightarrow$ ReLU $\rightarrow$
\texttt{Linear(in=256, out=256, bias=False)} $\rightarrow$ LN $\rightarrow$ ReLU
$\rightarrow$ \texttt{Linear(in=256, out=8, bias=False)} to decode $(x, y, z, l,
w, h, \sin(\theta), \cos(\theta))$. After the first attention block, we update
the query 3D position with the decoded 3D box, and re-sample the BEV features
based on the new 3D position to cross-attend to in the second attention block.
After the second attention block, we additionally use a lightweight MLP to
decode the proposal object class heatmap $\mathbf{c}\in\mathbb{R}^C$. The class
decoder is a sequence of \texttt{Linear(in=512, out=256, bias=False)}
$\rightarrow$ LN $\rightarrow$ ReLU $\rightarrow$ \texttt{Linear(in=256,
out=256, bias=False)} $\rightarrow$ LN $\rightarrow$ ReLU $\rightarrow$
\texttt{Linear(in=256, out=C, bias=False)} to output a logit for each of the $C$
classes. During training, we supervise both sets of the 3D proposal box
parameters, and the final class logits.

\paragraph{Multi-Modal Proposal Aggregation}

At the end of the multi-modal proposal stage, we first filter dense LiDAR proposals
as follows: we filter out detections with confidence $\le 0.01$
(confidence at BEV pixel $(i, j)$ is $\max_k{\mbox{Sigmoid}(h_{ijk})}$), apply
non-maximum suppression (NMS) with IoU threshold 0.2 in the 2D BEV space, and keep top 500 detections with the highest
confidence scores. Next, we aggregate both LiDAR and camera
3D proposals with BEV-space NMS with IoU threshold 0.2. In addition, in practice we found that
for \emph{nuScenes} the camera-proposal branch gives the best performance when
it is used to lift small and/or rare objects complementary to the LiDAR
proposals. As a result, in practice, we remove camera proposals for car,
trailer, truck, bus and construction vehicle classes before refinement.

\subsection{Refinement Stage}
In this subsection we provide more details with the transformer layers used in
object-LiDAR-camera attention blocks. First, as explained in the main paper, the
object query features $\in\mathbb{R}^{256}$ are initialized from either the
LiDAR features or the camera object queries. Both the deformable attention
blocks in LiDAR and camera cross-attention employ a linear layer with input dim
256 and output dim 2 to sample 2D offsets. The cross-attention transformer
layers have 8 attention heads with input dim 256, FFN feature dim 1024,
bias=False, and drop out=0.1. The self-attention transformer layer has 8
attention heads with input dim 256, FFN feature dim 256, bias=False, and drop
out=0.1.

After each object-camera-LiDAR attention block, we apply a lightweight box
decoder and a class decoder. The box decoder is a sequence of
\texttt{Linear(in=256, out=256, bias=False)} $\rightarrow$ LN $\rightarrow$ ReLU
$\rightarrow$ \texttt{Linear(in=256, out=256, bias=False)} $\rightarrow$ LN
$\rightarrow$ ReLU $\rightarrow$ \texttt{Linear(in=256, out=7, bias=False)} to
decode $(x, y, z, l, w, h, \theta)$ box parameters. The class decoder is a
sequence of \texttt{Linear(in=256, out=256, bias=False)} $\rightarrow$ LN
$\rightarrow$ ReLU $\rightarrow$ \texttt{Linear(in=256, out=C, bias=False)} to
decode logits for $C$ classes. During training we supervise both sets of 3D
boxes.
  \section{Additional Experiments Results}
\begin{table}
    \centering
    \scriptsize
    \setlength{\tabcolsep}{7pt}
    \begin{tabular}{lccccccccccc}\toprule
    $mAP_H$ & Method &Car &Adult &{\color{blue}Truck} &{\color{blue}CV} &{\color{blue}Bicycle} &{\color{blue}MC} &{\color{blue}Child} &{\color{blue}CW} &{\color{blue}Stroller} &{\color{blue}PP} \\\toprule
    \multirow{3}{*}{LCA0} &MMF~\cite{peri2022lt3d} &\ul{88.5} &86.6 &\ul{63.4} &29.0 &58.5 &68.2 &5.3 &35.8 &31.6 &39.3 \\
    &MMLF~\cite{ma2024mmlf} &86.3 &\ul{87.7} &60.6 &\ul{35.3} &\ul{70.0} &\ul{75.9} &\ul{8.8} &\ul{55.9} &\ul{37.7} &\textbf{58.1} \\
    &FOMO-3D &\textbf{88.9} &\textbf{90.7} &\textbf{65.1} &\textbf{36.6} &\textbf{72.8} &\textbf{80.2} &\textbf{29.8} &\textbf{60.2} &\textbf{40.1} &\ul{50.0} \\
    \midrule
    \multirow{3}{*}{LCA1} &MMF~\cite{peri2022lt3d} &\ul{89.4} &87.4 &\textbf{72.4} &31.3 &61.2 &69.7 &15.2 &52.0 &37.7 &39.4 \\
    &MMLF~\cite{ma2024mmlf} &86.8 &\ul{88.3} &68.5 &\ul{37.3} &\ul{70.4} &\ul{77.1} &16.2 &\ul{66.0} &\ul{51.5} &\textbf{58.2} \\
    &FOMO-3D &\ul{89.4} &\textbf{91.2} &\ul{70.3} &\textbf{39.2} &\textbf{73.9} &\textbf{81.5} &\textbf{59.2} &\textbf{73.4} &\textbf{54.8} &\ul{50.1} \\
    \midrule
    \multirow{3}{*}{LCA2} &MMF~\cite{peri2022lt3d} &\ul{89.5} &87.7 &\textbf{72.5} &31.5 &62.3 &69.9 &16.9 &56.3 &38.8 &39.8 \\
    &MMLF~\cite{ma2024mmlf} &86.9 &\ul{88.6} &68.6 &\ul{37.7} &\ul{70.9} &\ul{77.4} &16.3 &\ul{69.0} &\ul{52.4} &\textbf{58.9} \\
    &FOMO-3D &\ul{89.5} &\textbf{91.5} &\ul{70.4} &\textbf{39.7} &\textbf{74.9} &\textbf{81.9} &\textbf{59.5} &\textbf{77.1} &\textbf{58.2} &\ul{50.6} \\

    \bottomrule
    \end{tabular}
    \caption{\textbf{[nuScenes] Class-specific hierarchical metrics}. CV = Construction Vehicle. MC = Motorcycle. CW = Construction Worker. PP = Pushable-Pullable. \texttt{Medium} and \texttt{Few} classes are in \color{blue}{blue}.}
    \label{tab:nusc-lca}
    \end{table}
\paragraph{\emph{nuScenes} results with hierarchical mAP.} Following previous
works~\cite{peri2022lt3d,ma2024mmlf}, we also adopt the hierarchical mAPs
($mAP_H$), which reports mAP with three tiers based on the least common ancestor
(LCA) distance: LCA0 is the standard per-class mAP, LCA1 treats each object
class as one of the three parent classes (vehicle, pedestrian, movable object)
and tolerates misclassification with sibling classes, and LCA2 measures
class-agnostic mAP on \emph{nuScenes}. Table~\ref{tab:nusc-lca} showcases
per-class hierarchical metrics, comparing ours against previous LT3D
methods~\cite{peri2022lt3d,ma2024mmlf}. FOMO-3D improves the hierarchical
metrics for almost all classes in all three LCA tiers, showing that FOMO-3D is
not only better at fine-grained classification for almost all object classes at
LCA0, but also improves the general localization and detection quality at LCA1
and LCA2 as well.

\begin{table}[!htp]\centering
    \scriptsize
    \setlength{\tabcolsep}{3pt}
    \begin{tabular}{ccccc}\toprule
        Frustum Attention &\texttt{All} &\texttt{Many} &\texttt{Medium} &\texttt{Few} \\\midrule
        & 53.6 &80.9 &58.6 &25.1 \\
        \checkmark &54.6 &79.9 &59.6 &27.6 \\
        \bottomrule
        \end{tabular}
    \caption{\textbf{[nuScenes]} Frustum attention ablation.}
\label{tab:nusc-frust-attn-ablation}
\end{table}
\begin{table*}\centering
    \scriptsize
    \begin{tabular}{ccccccc}\toprule
        Frustum Attention &Vehicle &Towed &Cone &Person &Cyclist \\\midrule
         &87.0 &73.6 &86.9  &64.1 &76.4 \\
        \checkmark  &87.2 &74.2 &88.9 &68.7 &79.3 \\
        \bottomrule
        \end{tabular}
    \caption{\textbf{[Highway]} Frustum attention ablation.}
\label{tab:waabi-ablation-frust-attn}
\end{table*}
\paragraph{Effect of frustum attention.} In the camera-based proposal branch, we
initialize object queries with 2D OWL detection bounding boxes and associated
OWL features and M3D depths, refine the features with frustum-based attention,
and decode 3D bounding boxes as camera-based proposals. Here we ablate the
effect of frustum attention with an experiment where the camera-based proposal
branch learns to decode a 3D box directly from the initial query features.
Table~\ref{tab:nusc-frust-attn-ablation} and
Table~\ref{tab:waabi-ablation-frust-attn} show results on the \emph{nuScenes}
and \emph{Highway} datasets respectively and demonstrate that overall frustum
attention is effective, especially on rare classes and in the long-range
\emph{Highway} setting with more apparent depth errors.

\begin{table}[!htp]\centering
    \scriptsize
    \setlength{\tabcolsep}{3pt}
    \begin{tabular}{cccccccc}\toprule
        $N_x$ & $N_y$ & $N_z$ & $\delta$ &\texttt{All} &\texttt{Many} &\texttt{Medium} &\texttt{Few} \\\midrule
        0 & 0 & 20 & 10 &53.5 & 79.8 & 58.2 & 26.1\\
        3 & 3 & 20 & 10 &55.0 & 80.2 & 59.3 & 29.0\\
        1 & 1 & 10 & 10 &54.6 & 79.9 & 59.1 & 28.2\\
        1 & 1 & 40 & 10 &54.3 & 79.8 & 58.8 & 27.9\\
        1 & 1 & 10 & 4 &53.8 & 79.6 & 58.8 & 26.3\\
        1 & 1 & 10 & 25 & 52.6 & 79.9 & 59.1 & 22.8\\\midrule
        1 & 1 & 20 & 10 &54.6 &79.9 &59.6 &27.6 \\
        \bottomrule
        \end{tabular}
    \caption{\textbf{[nuScenes]} Sampling resolution ablation.}
\label{tab:nusc-sample-grid-ablation}
\end{table}
\paragraph{Effect of frustum attention sampling grid resolution.} In frustum
attention, we construct a mesh grid to sample from frustum features for
computational efficiency. Table~\ref{tab:nusc-sample-grid-ablation} showcases
how various combinations of $(N_x, N_y, N_z, \delta)$ affect the final
performance. Better performance is achieved with more sampling rays
(\emph{i.e.}, bigger $N_x$ and $N_y$) but with heavier computation costs. We
chose the hyperparameters in the final row to balance computation and
performance.

\begin{table}[!htp]\centering
    \scriptsize
    \begin{tabular}{lccccc}\toprule
        Method & Frustum-based Loss Constraint &\texttt{All} &\texttt{Many} &\texttt{Medium} &\texttt{Few} \\\midrule
        $M_1$ &  & 54.0 &80.3 &59.5 &25.8 \\
        $M_2$ (FOMO-3D) & \checkmark &\textbf{54.6} &79.9 &59.6 &\textbf{27.6} \\
        \bottomrule
        \end{tabular}
    \caption{\textbf{[nuScenes]} Ablation for frustum-based matching constraint in camera proposals.}
\label{tab:nusc-rarity-ablation-frustum-matching}
\end{table}
\paragraph{Effect of frustum-based matching loss.} When training the camera
proposals, we add a hard constraint in detection to ground-truth label matching
that the ground-truth label must be present in the detection frustum space. We
perform an ablation on the \emph{nuScenes} dataset to compare with using regular
IoU-based matching without the frustum-based hard constraint.
Table~\ref{tab:nusc-rarity-ablation-frustum-matching} shows that with this hard
constraint ($M_1 \rightarrow M_2$), the mAP for \texttt{Few} is two points
better.

\begin{table}[!htp]\centering
    \scriptsize
    \setlength{\tabcolsep}{3pt}
    \begin{tabular}{cccccc}\toprule
        Method & Depth Estimator &\texttt{All} &\texttt{Many} &\texttt{Medium} &\texttt{Few} \\\midrule
        $M_1$ & LiDAR  &  53.0 & 79.9 & 58.9 & 23.7 \\
        $M_2$ & M3D Rescaled  & 53.5 & 79.9 & 58.9 & 25.2 \\
        $M_3$ (FOMO-3D) & M3D  & 54.6 & 79.9 & 59.6 & 27.6 \\
        \bottomrule
        \end{tabular}
    \caption{\textbf{[nuScenes]} LiDAR depth ablations.}
\label{tab:nusc-depth-ablation}
\end{table}
\paragraph{Alternative depths.} To understand the effect of using M3D depths, we
ablate (1) replacing M3D with (sparse) LiDAR depths, and (2) rescaling M3D
depths with LiDAR depths with bucketized median scaling at near (0m-10m), medium
(10-30m), and far (30m+) ranges. Table~\ref{tab:nusc-depth-ablation} shows that
using dense M3D depths directly on the \emph{nuScenes} dataset has the best
results.

\begin{table}[!htp]\centering
    \scriptsize
    \setlength{\tabcolsep}{3pt}
    \begin{tabular}{ccccccc}\toprule
        Camera & LiDAR & Object &\texttt{All} &\texttt{Many} &\texttt{Medium} &\texttt{Few} \\\midrule
        & \checkmark & \checkmark & 41.0 &76.1 &45.3 &6.7 \\
        \checkmark & & \checkmark &52.0 &80.0 &58.3 &23.4 \\
        \checkmark & \checkmark & & 54.1 & 81.0 &60.1 & 24.8 \\
        \checkmark & \checkmark & \checkmark &54.6 &79.9 &59.6 &27.6 \\
        \bottomrule
        \end{tabular}
    \caption{\textbf{[nuScenes]} Refinement type ablations.}
\label{tab:nusc-refinement-ablation}
\end{table}
\paragraph{Effect of refinement modules.} To understand how each type of
refinement attention in the refinement stage contributes to the final
performance, Table~\ref{tab:nusc-refinement-ablation} ablates each attention
type in the \emph{nuScenes} dataset, and shows that all of them are key to
maximizing performance on long-tailed detection.

\begin{table*}\centering
    \scriptsize
    \begin{tabular}{lcccccccc}\toprule
        Method & Prop & Refine &Vehicle &Towed &Cone &Person &Cyclist \\\midrule
        $M_1$ & L & L  &86.7 &73.5 &82.6  &58.4 &74.8 \\
        $M_2$ & L & L+C  &\ul{86.9} &\ul{73.8} &\ul{86.0} &\ul{65.4} &\ul{77.7} \\
        $M_3$ & L+C & L+C &\textbf{87.2} &\textbf{74.2} &\textbf{88.9} &\textbf{68.7} &\textbf{79.3} \\
        \bottomrule
        \end{tabular}
    \caption{\textbf{[Highway]} Ablations.}
\label{tab:waabi-ablation}
\end{table*}

\paragraph{Ablation table on \emph{Highway}.} In the main paper, the ablation
results for the \emph{Highway} dataset is presented in terms of net gains over
the LiDAR-only model in Fig. 5. For completeness, Table~\ref{tab:waabi-ablation}
shows the detailed numerical metrics for the lidar-only model, lidar-proposal +
multi-modal refinement, and multi-modal proposal + multi-modal refinement
respectively. The results show that both camera proposal and camera attention
during refinement are helpful.

\begin{table}[!htp]\centering
    \scriptsize
    \begin{tabular}{lcccccc}\toprule
        Method & Proposal Stage Fusion & Learned Depth &\texttt{All} &\texttt{Many} &\texttt{Medium} &\texttt{Few} \\\midrule
        $M_1$ & Feature-level & & 54.8 &80.2 &60.6 &26.9 \\
        $M_2$ & Feature-level & \checkmark &53.3 &79.6 &59.0 &24.7 \\
        $M_3$ (FOMO-3D) & Late &  &54.6 &79.9 &59.6 &27.6 \\
        \bottomrule
        \end{tabular}
    \caption{\textbf{[nuScenes]} Explorations for alternative proposal-level fusion strategies.}
\label{tab:nusc-rarity-ablation-fusion}
\end{table}
\begin{table*}\centering
    \scriptsize
    \begin{tabular}{lcccccccc}\toprule
        Method & Proposal Stage Fusion & Learned Depth &Vehicle &Towed &Cone &Person &Cyclist \\\midrule
        $M_1$ & Feature-level &  &86.1 &71.7 &86.5  &64.6 &76.4 \\
        $M_2$ & Feature-level & \checkmark  &87.0 &73.7 &87.0 &64.2 &74.7 \\
        $M_3$ (FOMO-3D) & Late &  &\textbf{87.2} &\textbf{74.2} &\textbf{88.9} &\textbf{68.7} &\textbf{79.3} \\
        \bottomrule
        \end{tabular}
    \caption{\textbf{[Highway]} Explorations for alternative proposal-level fusion strategies.}
\label{tab:waabi-ablation-fusion}
\end{table*}
\paragraph{Alternative fusion strategies.} In the proposal stage of FOMO-3D, we
adopt a late-fusion design where we merge the LiDAR-based 3D proposals and
camera-based 3D proposals in the end. Alternatively, due to the effectiveness of
feature-level fusion in BEVFusion~\cite{liu2022bevfusion}, it is also worth
exploring feature-level fusion with OWL features in the proposal stage. We
therefore designed two alternative proposal-level fusion strategies for
exploration.

In the first method, we replace the proposal stage with a dense feature lifting
design, where we lift every OWL token in the image with respective M3D depths,
fuse with the LiDAR features with concatenation, and decode a single set of 3D
proposals to be refined later. This fusion strategy is similar to BEVFusion,
except we are only using a single depth from M3D instead of learning and lifting
with a depth distribution.

In the second method, instead of lifting with the M3D depth directly, we use a
learned depth based on M3D. We first create 20 scale factors that are based on
uniform intervals in [0.5, 1.5), \emph{i.e.}, $\mathbf{s} = [0.5, 0.55, 0.6,
\ldots, 1.45] \in \mathbb{R}^{20}$. Each scale factor will be multiplied with
the original M3D depth to create 20 depth buckets. We then apply a lightweight
MLP on each OWL token to learn $\mathbf{p} \in \mathbb{R}^{20}$, representing
the confidence of each depth bucket. We finally aggregate the learned depth with
$d' = \frac{1}{20}\sum_{i=0}^{19}{s_i\cdot p_i \cdot d}$ where $d$ is the
original M3D depth, and $p_i$ is the respective confidence in
$\mbox{Softmax}(\mathbf{p})$. We lift every OWL token with the learned depth,
fuse with the LiDAR features similar to the first method and decode a single set
of proposals. Compared to the first method, the second method has more
flexibility to correct depth errors. Unlike BEVFusion, we do not lift the same
token with multiple depths due to memory constraints.

Table~\ref{tab:nusc-rarity-ablation-fusion} shows the comparison for these two
alternative fusion methods against FOMO-3D on the \emph{nuScenes} dataset.
Interestingly, the feature-level fusion method that uses M3D depth directly
($M_1$) has comparable performance with FOMO-3D, thanks to M3D's accurate
zero-shot depths in the short evaluation range of
\emph{nuScenes}~\cite{hu2024metric3dv2}. The learned depth version ($M_2$) is
slightly worse, possibly due to the fact that the learned depth introduces more
noises to the already accurate M3D initializations. In addition, it's worth
noting that $M_1$ has noticeably better performance on the \texttt{Medium}
group, which implies that directly using all features without any prompting
combined with feature-level fusion could be very effective at exploiting 2D
foundation models when the depth is very accurate. However, on the
\emph{Highway} dataset with challenging longer evaluation ranges, M3D has more
errors especially in the far range. As a result, naively lifting and fusing with
M3D depths results in worse performance for $M_1$ in
Table~\ref{tab:waabi-ablation-fusion}, and learned depths over depth buckets
$M_2$ is overall better than $M_1$. FOMO-3D outperforms both these baselines,
which shows that FOMO-3D's camera proposal branch with more elaborate 3D lifting
and attention-based multi-modal fusion strategies is more robust to M3D depth
errors and more effective at lifting 2D foundation model priors to 3D.

\paragraph{Latency and Memory.}
Benchmarking nuScenes inference on an RTX A5000 GPU yields 79.8ms for lidar
proposal (CenterPoint), 97.7ms for camera proposal (\textbf{excluding foundation
model inference}), and 131.2ms for refinement, with 18G memory usage. This is
comparable with 80.7ms for CenterPoint, 321.4ms for FUTR3D, 119.2ms for
BEVFusion reported in the BEVFusion paper~\cite{liu2022bevfusion}. These modules
can be optimized with custom CUDA kernels to reduce the gap with \emph{e.g.}
BEVFusion, but real-time inference is beyond the scope of this work. Based on
the official implementation of M3D-Giant and HuggingFace implementation of
OWLv2-Large, inference time is 1590 ms per image for M3D and 834 ms per image
crop ($960\times 960$ pixels) for OWL. During training and evaluation, we cache
M3D/OWL outputs as a one-time pre-processing step, so these costs are
negligible. As addressed in the limitations section, improving run time for
real-time applications is an exciting direction for future research
(\emph{e.g.}, distillation into a smaller model).

\begin{figure*}
    \hspace*{-1.4in}
    \includegraphics[width=1.5\linewidth]{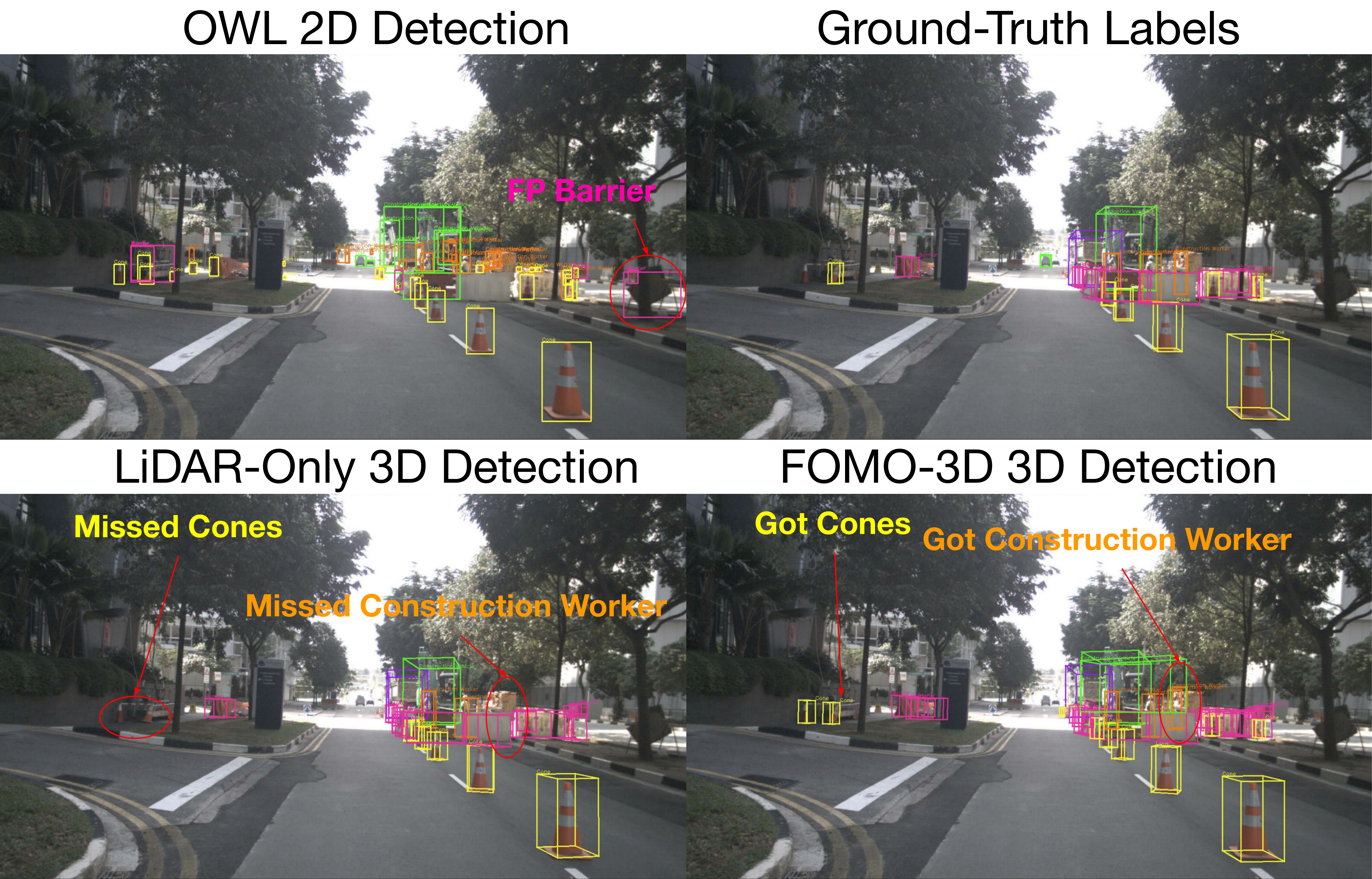} 
    \caption{\textbf{[nuScenes]} Qualitative example \#1. In this example, we
    added OWL prompts for barriers to illustrate low-quality zero-shot barrier
    detections. As shown in the visualization, OWL produces many false positive
    barriers without capturing most of the true positive barriers. As a result
    we removed barrier prompting from the camera proposal branch. In addition,
    the LiDAR-only model misses small objects such as cones and construction
    workers, but OWL and FOMO-3D are able to detect them successfully.}
    \label{fig:nuscenes-qual-1}   
\end{figure*}

\begin{figure*}
    \hspace*{-1.4in}
    \includegraphics[width=1.5\linewidth]{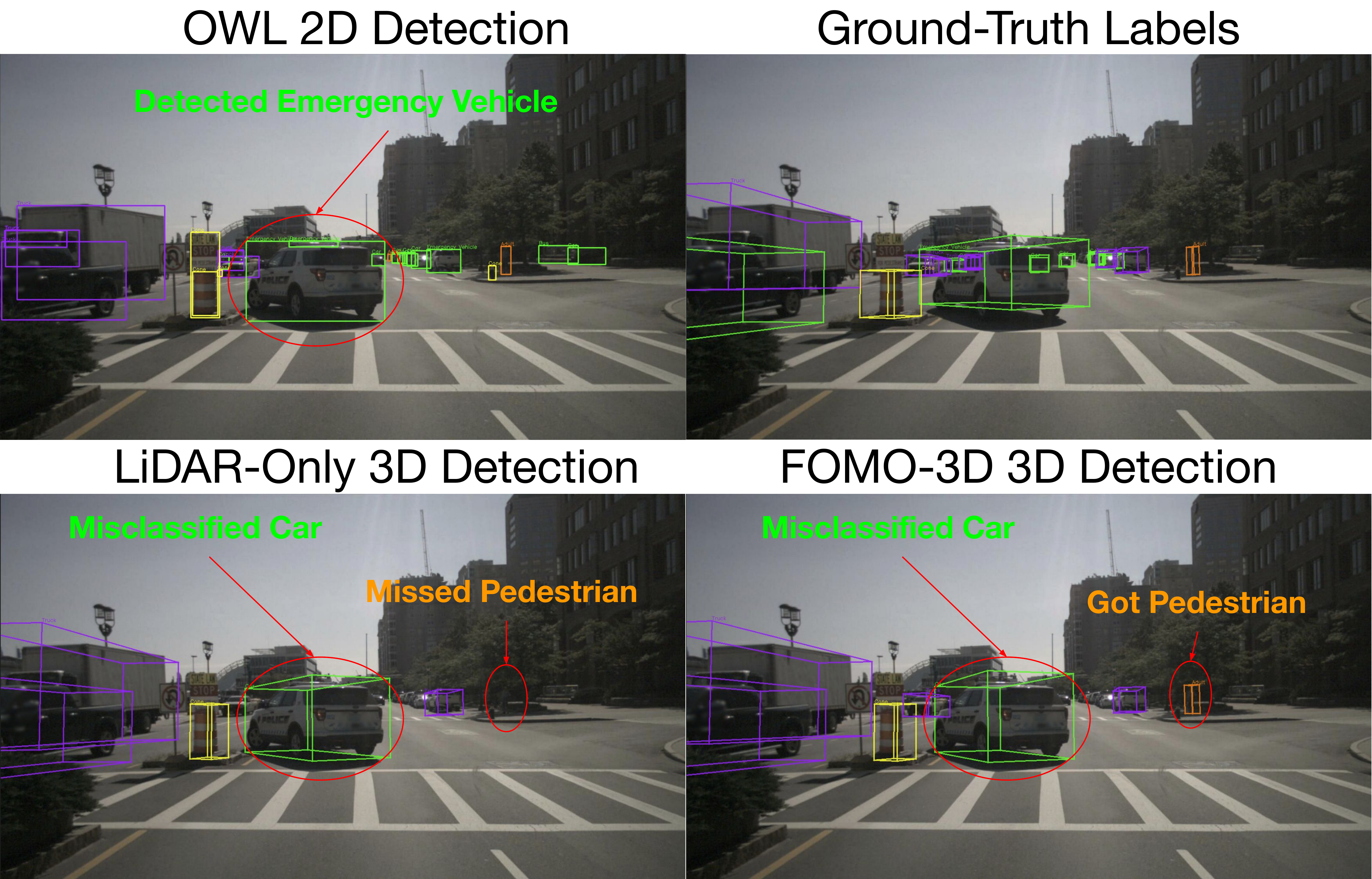} 
    \caption{\textbf{[nuScenes]} Qualitative example \#2. In this example, OWL
    correctly detects and classifies an adult and an emergency vehicle. The
    LiDAR-only model misses the adult (because it is small and sparsely observed
    by LiDAR) and misclassifies the emergency vehicle as car due to lack of
    semantics. FOMO-3D is able to successfully detect the adult thanks to OWL,
    but fails to retain the original OWL emergency vehicle classification.}
    \label{fig:nuscenes-qual-2}   
\end{figure*}

\begin{figure*}
    \hspace*{-1.4in}
    \includegraphics[width=1.5\linewidth]{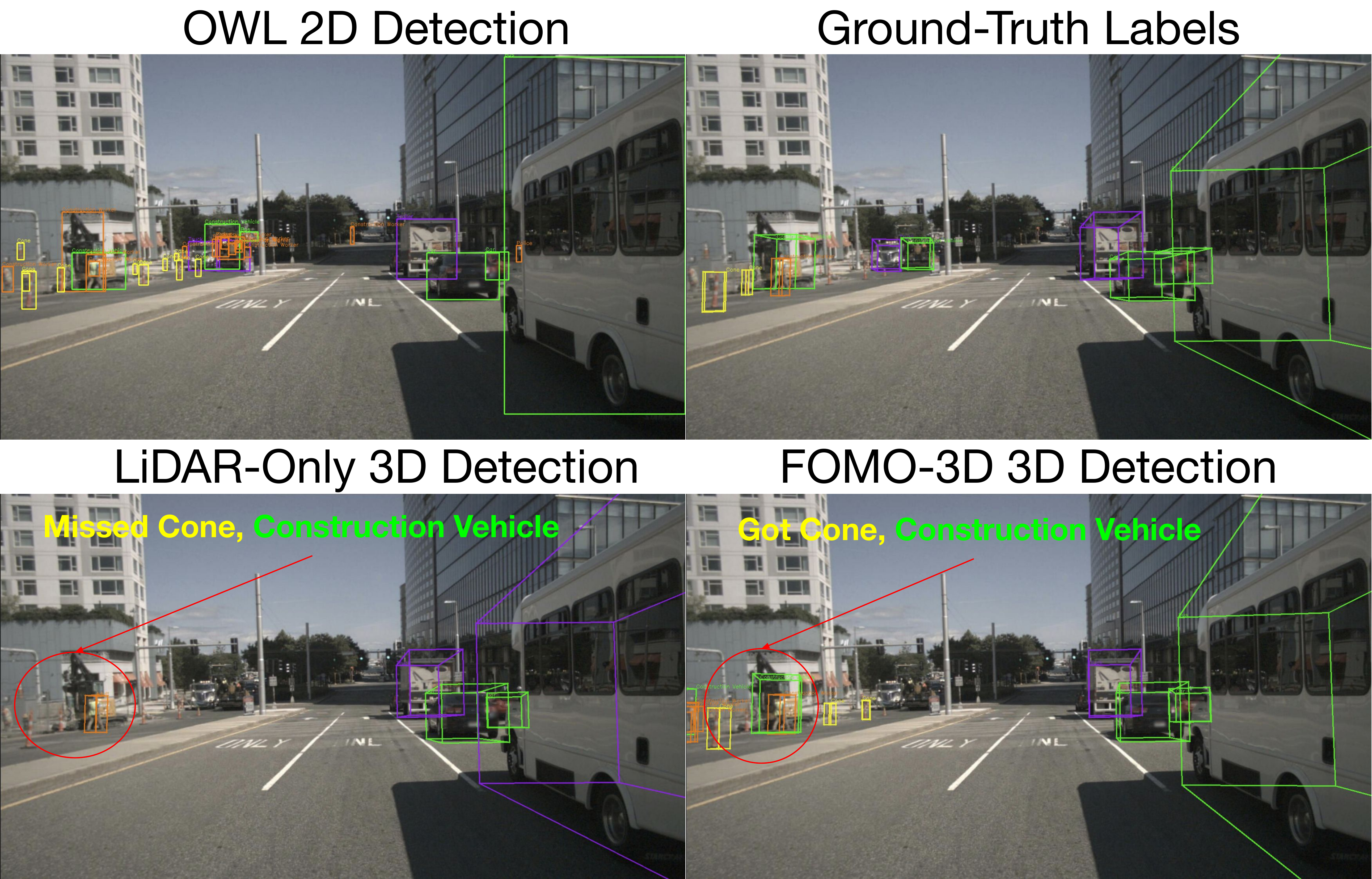} 
    \caption{\textbf{[nuScenes]} Qualitative example \#3. In this example, the
    LiDAR-only model fails to detect a cone and a construction vehicle, while
    FOMO-3D is able to detect and classify them successfully.}
    \label{fig:nuscenes-qual-3}   
\end{figure*}

\begin{figure*}
    \hspace*{-1.4in}
    \includegraphics[width=1.5\linewidth]{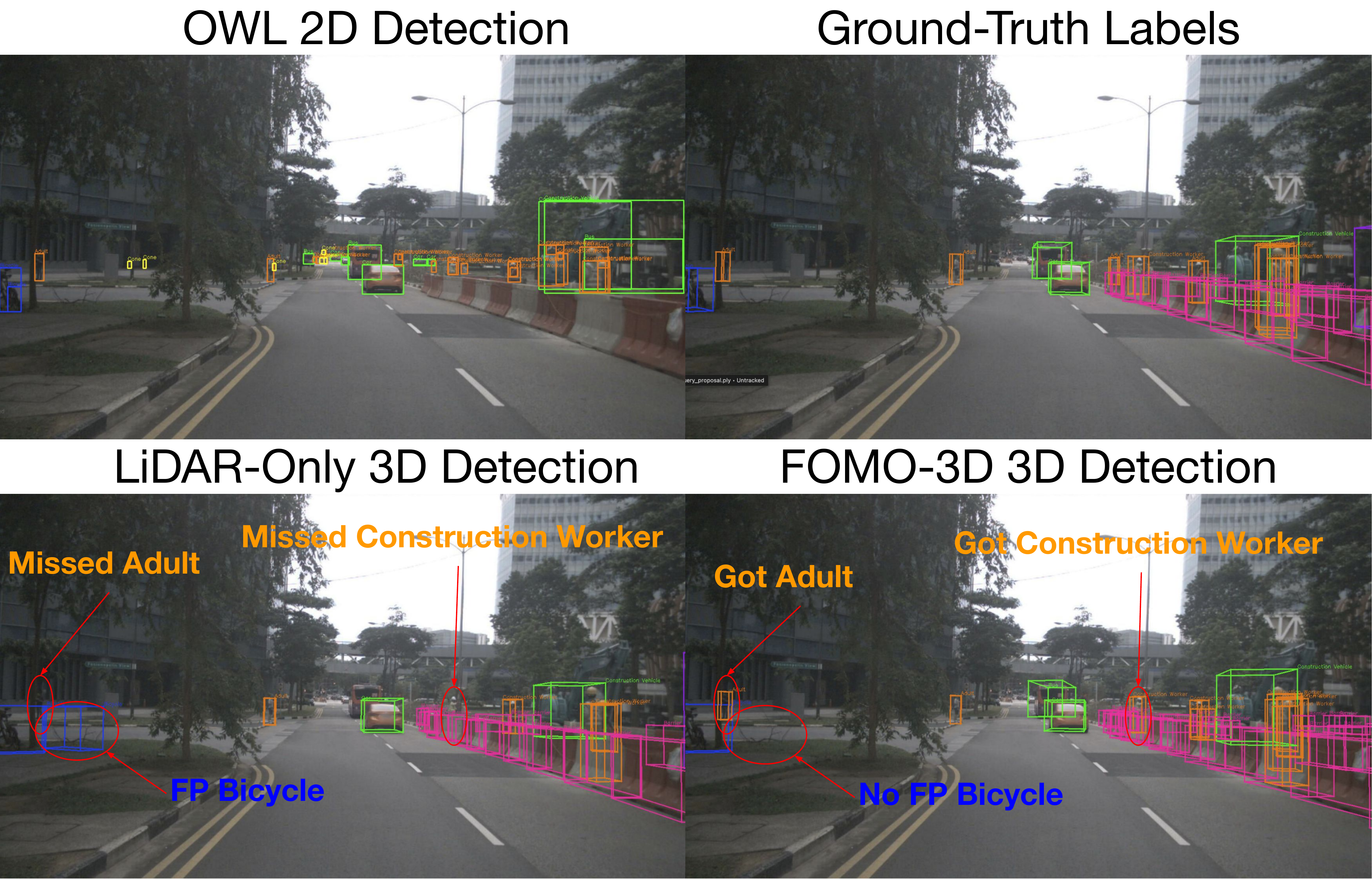} 
    \caption{\textbf{[nuScenes]} Qualitative example \#4. In this example, the
    LiDAR-only model fails to detect an adult and a construction worker and also
    outputs a false positive bicycle, while FOMO-3D is able to detect them
    successfully while rejecting the false positive.}
    \label{fig:nuscenes-qual-4}   
\end{figure*}

\begin{figure*}
    \hspace*{-1.4in}
    \includegraphics[width=1.5\linewidth]{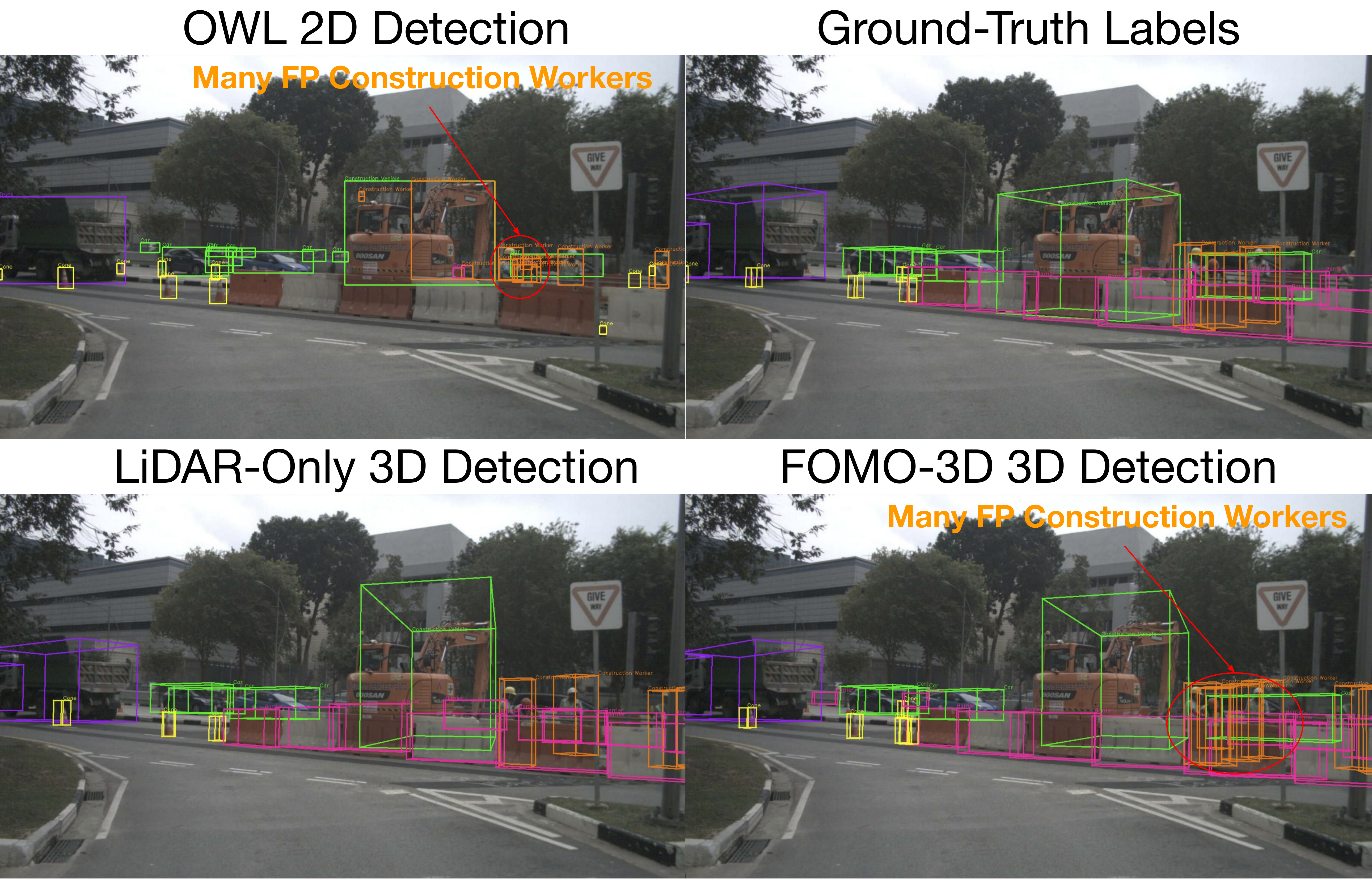} 
    \caption{\textbf{[nuScenes]} Qualitative example \#5. This example shows a
    failure case of OWL and FOMO-3D. OWL produces many duplicated construction
    workers. While FOMO-3D lifts them to the correct construction worker size,
    it fails to reject some duplicated false positives in the final detections.
    This is a challenging case because these false positives are adjacent to
    true positives in the image. False positives can be generated in the camera
    proposal stage directly from OWL proposals. In the refinement stage, many
    proposals which project onto the same image region could pick up features
    describing construction workers, and multiple proposals can attend to the
    same image features to generate these false positives. Designing a more
    optimal false positive and duplicate removal method is a future direction to
    improve FOMO-3D.}
    \label{fig:nuscenes-qual-5}   
\end{figure*}

\paragraph{Additional qualitative results for \emph{nuScenes}.}
Fig.~\ref{fig:nuscenes-qual-1}, ~\ref{fig:nuscenes-qual-2},
~\ref{fig:nuscenes-qual-3}, ~\ref{fig:nuscenes-qual-4}
and~\ref{fig:nuscenes-qual-5} provide more qualitative examples on the
\emph{nuScenes} dataset. In general, the LiDAR-only model often misses small
and/or rare objects such as cone, adult and construction worker. OWL has great
zero-shot detections for these small objects, but can have false positives
especially when there are true positives in the local region. FOMO-3D is able to
recover small and/or rare objects in most cases, with some misclassification and
false positive failure cases discussed in Fig.~\ref{fig:nuscenes-qual-2} and
Fig.~\ref{fig:nuscenes-qual-5}.

\begin{figure*}
    \hspace*{-1.4in}
    \includegraphics[width=1.5\linewidth]{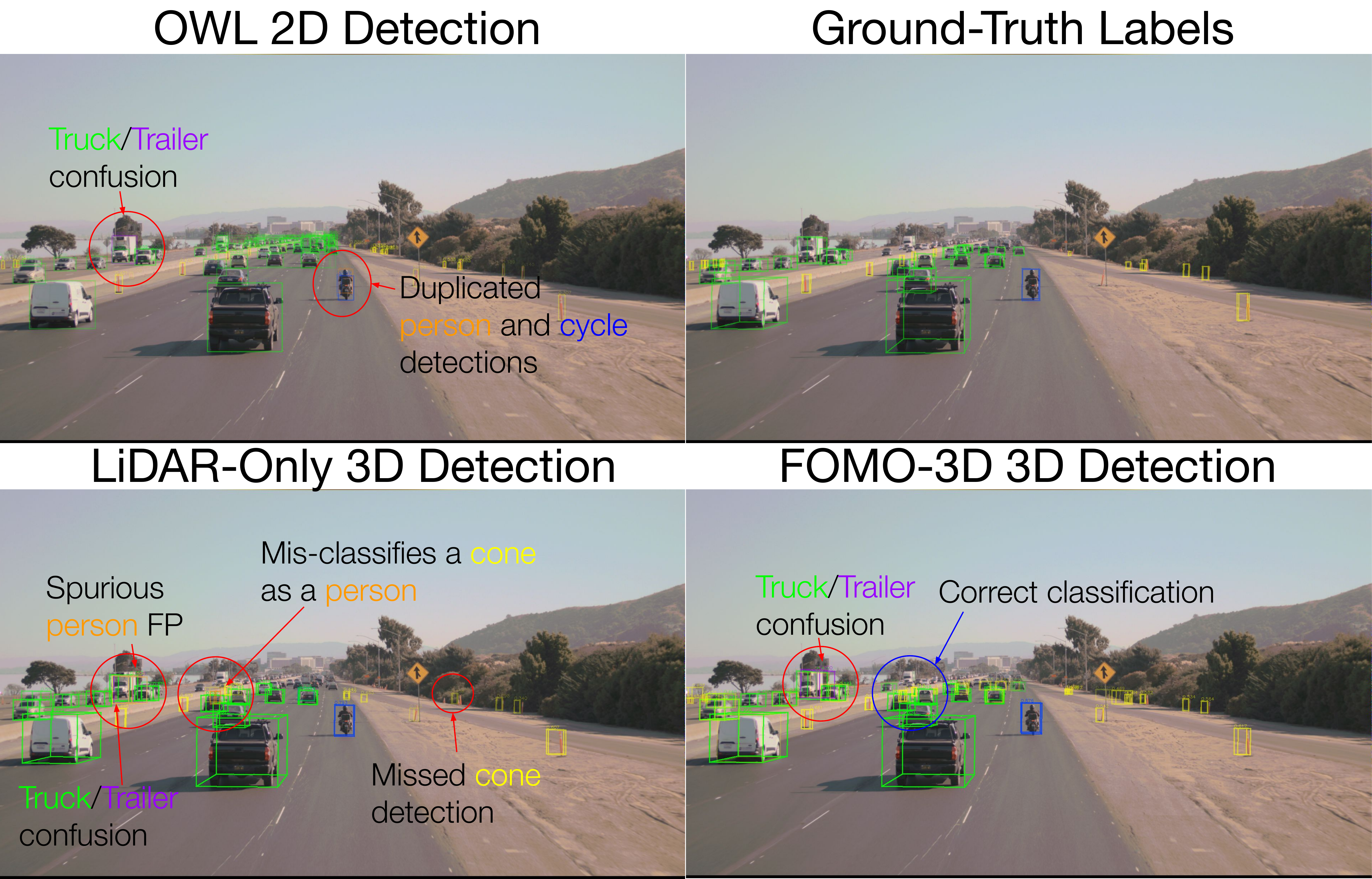} 
    \caption{\textbf{[Highway]} Qualitative example \#1. {\color{green} Green}
    denotes vehicle. {\color{purple} Purple} denotes towed object. {\color{blue}
    Blue} denotes cyclists. {\color{orange} Orange} denotes person.
    {\color{yellow} Yellow} denotes cone. Opacity of the boxes reflects
    detection confidence (higher confidence corresponds to more solid lines).
    FP=False Positive. FOMO-3D is able to correct various errors in OWL and
    LiDAR-only detections, but still mis-classifies a truck as trailer.}
    \label{fig:highway-qual-1}   
\end{figure*}

\begin{figure*}
    \hspace*{-1.4in}
    \includegraphics[width=1.5\linewidth]{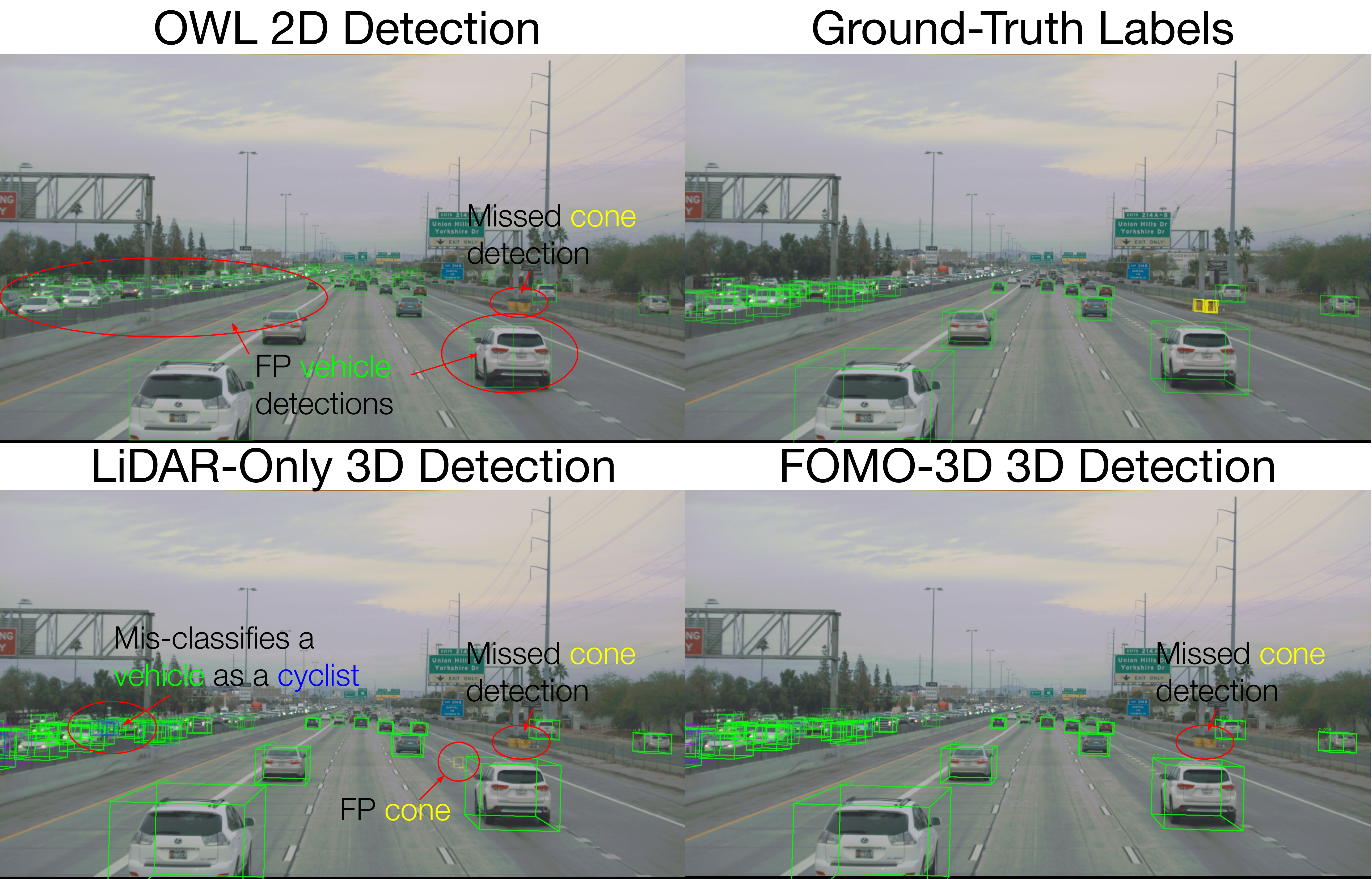} 
    \caption{\textbf{[Highway]} Qualitative example \#2. {\color{green} Green}
    denotes vehicle. {\color{purple} Purple} denotes towed object. {\color{blue}
    Blue} denotes cyclists. {\color{orange} Orange} denotes person.
    {\color{yellow} Yellow} denotes cone. Opacity of the boxes reflects
    detection confidence (higher confidence corresponds to more solid lines).
    FP=False Positive. This example shows a very dense traffic scene on the
    highway. OWL has impressive zero-shot 2D detections, but it also outputs a
    few false positives. LiDAR-only model mis-classifies a vehicle as a cyclist,
    and also draws a false positive cone. FOMO-3D is able to correctly classify
    in most cases without having many false positives. All three detectors miss
    two cones on the right.}
    \label{fig:highway-qual-2}   
\end{figure*}

\begin{figure*}
    \hspace*{-1.4in}
    \includegraphics[width=1.5\linewidth]{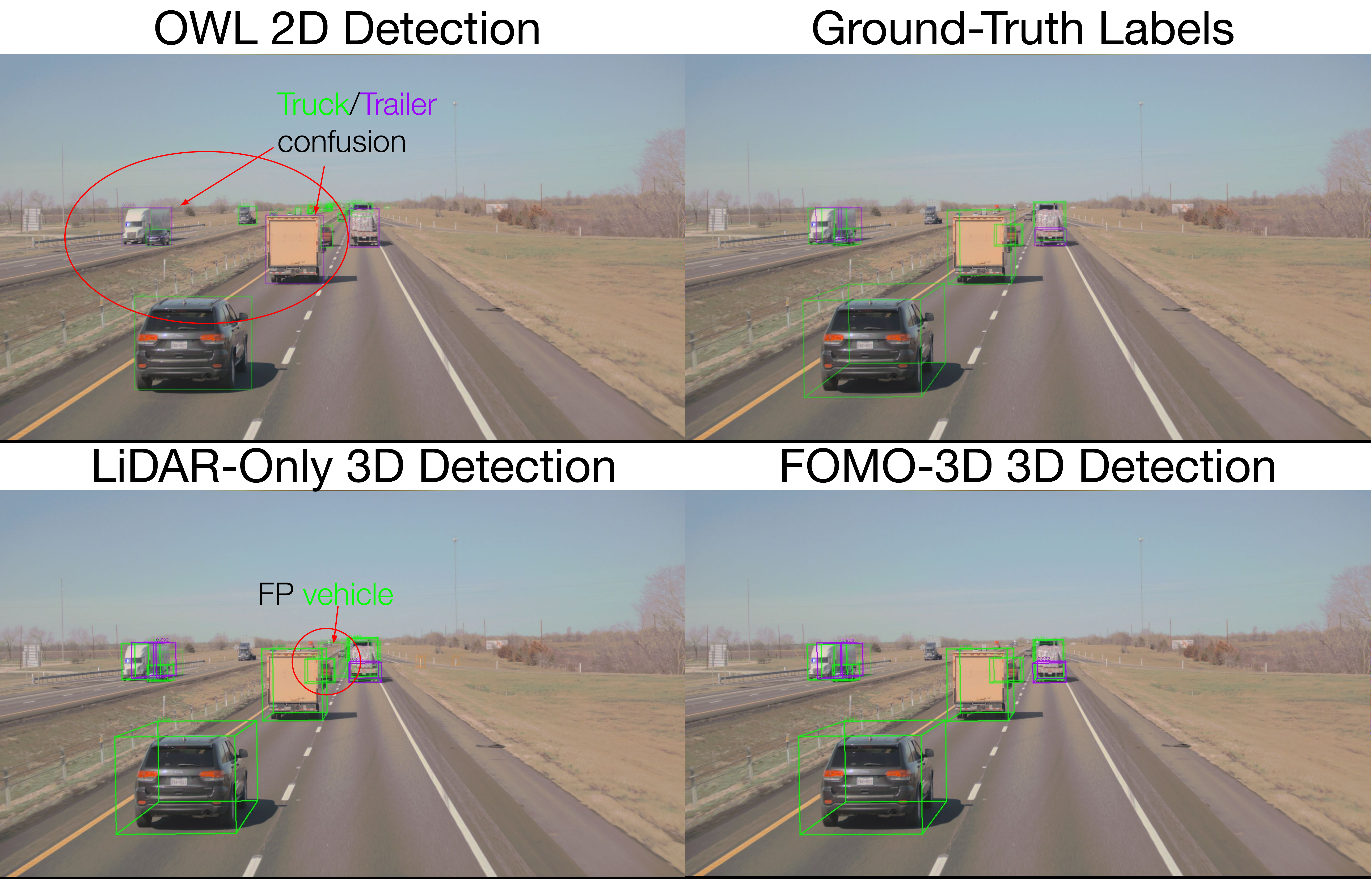} 
    \caption{\textbf{[Highway]} Qualitative example \#3. {\color{green} Green}
    denotes vehicle. {\color{purple} Purple} denotes towed object. {\color{blue}
    Blue} denotes cyclists. {\color{orange} Orange} denotes person.
    {\color{yellow} Yellow} denotes cone. Opacity of the boxes reflects
    detection confidence (higher confidence corresponds to more solid lines).
    FP=False Positive. Unlike OWL, FOMO-3D is able to utilize 3D LiDAR
    information and the training data to distinguish between vehicles and towed
    objects. FOMO-3D is able to effectively fuse 3D and 2D information and learn
    from the training data. It also outputs fewer false positives compared to
    the LiDAR-only model.}
    \label{fig:highway-qual-3}   
\end{figure*}

\begin{figure*}
    \hspace*{-1.4in}
    \includegraphics[width=1.5\linewidth]{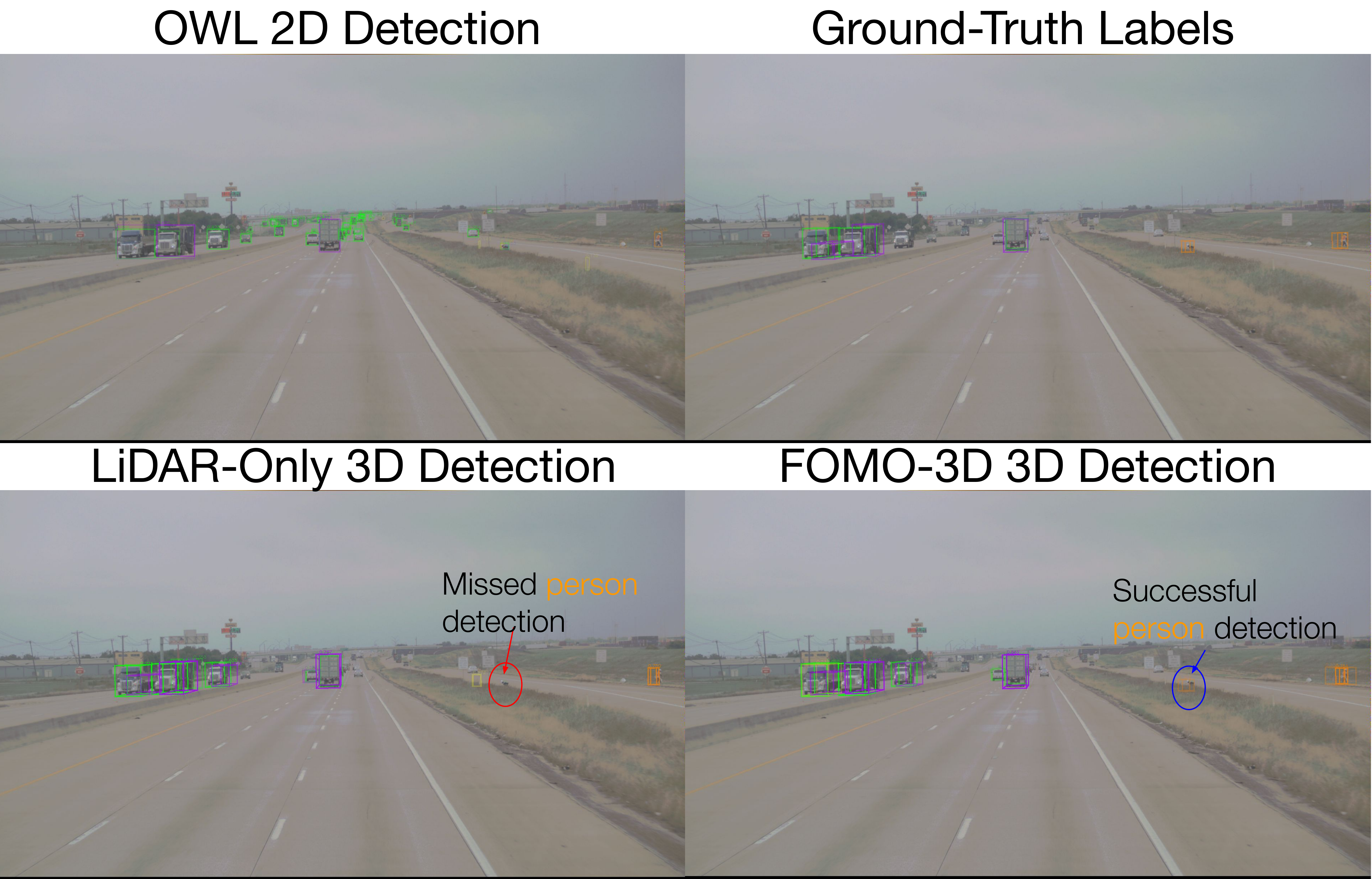} 
    \caption{\textbf{[Highway]} Qualitative example \#4. {\color{green} Green}
    denotes vehicle. {\color{purple} Purple} denotes towed object. {\color{blue}
    Blue} denotes cyclists. {\color{orange} Orange} denotes person.
    {\color{yellow} Yellow} denotes cone. Opacity of the boxes reflects
    detection confidence (higher confidence corresponds to more solid lines).
    FP=False Positive. In this example, the LiDAR-only model misses the person
    on the right, while OWL is able to detect it. With effective multi-modal
    fusion, FOMO-3D successfully detects the person.}
    \label{fig:highway-qual-4}   
\end{figure*}

\paragraph{Additional qualitative results for \emph{Highway}.}
Fig.~\ref{fig:highway-qual-1}, ~\ref{fig:highway-qual-2},
~\ref{fig:highway-qual-3} and~\ref{fig:highway-qual-4} provide qualitative
examples on the \emph{Highway} dataset. In general, although OWL has impressive
zero-shot 2D detection boxes, it oftentimes outputs false positives and confuses
vehicle and towed object classes. LiDAR-only detections learn to distinguish
between vehicles and towed objects in most cases, but confuse cone with person
(due to their similar cylindrical geometries and smaller sizes) and tends to
output false positives around spurious LiDAR points. FOMO-3D is able to leverage
both 2D and 3D information for correct detection and classification. For
truck/trailer confusion, FOMO-3D shares similar failure cases as the LiDAR-only
model.

\end{document}